\documentclass[11pt]{article}

\usepackage[preprint]{acl}

\usepackage{times}
\usepackage{latexsym}

\usepackage[T1]{fontenc}

\usepackage[utf8]{inputenc}
\usepackage{microtype}
\usepackage{inconsolata}
\usepackage{graphicx}
\usepackage{amsmath}
\usepackage{booktabs}
\usepackage{subcaption}
\usepackage{tabularx}
\usepackage{cleveref}
\usepackage[labelfont=bf]{caption}
\usepackage{subcaption}
\usepackage{fvextra}
\usepackage{hyperref}
\usepackage{array}

\DefineVerbatimEnvironment{PromptBlock}{Verbatim}{
    fontsize=\footnotesize,
    breaklines=true,
    breakanywhere=true,
    breaksymbolleft={},
    breaksymbolright={},
    frame=single,
    framesep=2mm
}

\title{How Post-Training Influences Verbalized Algorithmic Computation in Language Models}

\author{Niall Dalton \\
  University College London \\
  \texttt{niall.dalton.25@ucl.ac.uk}}

\begin{document}

\maketitle

\begin{abstract}
Post-training improves language-model reasoning, but how it changes the use of internal information and verbalized chain of thought (CoT) remains unclear. We study successive SFT, DPO, and RLVR checkpoints of Olmo-3-7B-Instruct on multi-digit addition, using carry values as known intermediate state. Probes recover carry information from internal activations with similar accuracy in SFT and RLVR, although manipulating this information does not reliably control model outputs. In contrast, the usefulness of verbalized reasoning differs: CoTs produced by RLVR yield higher accuracy across receiving checkpoints than those produced by SFT or DPO. We also show that carry-related statements receive higher importance scores after RLVR than matched random controls. DPO shows similar sensitivity to carry-related statements but without comparable accuracy gains. These results suggest that post-training changes the usefulness of written reasoning and sensitivity to its content in meaningful ways, although greater sensitivity does not necessarily coincide with better problem performance.
\end{abstract}

\section{Introduction}
Large Reasoning Models (LRMs) have introduced a paradigm shift in language model development and achieved breakthrough performance in many areas \citep{Guo_2025_deepseek_r1}. Despite this, the functional role and necessity of reasoning through the Chain-of-Thought (CoT) is unclear, with some research calling into question the nature of such reasoning, and whether LRMs are using explicit algorithms or some form of heuristics \citep{shojaee2025illusionthinkingunderstandingstrengths, daliberti2026illusioninsightreasoningmodels, boppana2026reasoningtheaterdisentanglingmodel}.

In particular, Reinforcement Learning from Verifiable Rewards (RLVR) has been attributed to the success of LRMs, indicating the importance of post-training \citep{yu2025dapoopensourcellmreinforcement, Guo_2025_deepseek_r1}. However, \citet{venhoff2025basemodelsknowreason} provide strong evidence that base models often contain many of the same mechanisms as their LRM counterparts have and, rather than teaching a model these mechanisms, RL based post-training encourages the use of them in a more productive way (that is, to solve the particular problem at hand correctly). 

While conventional algorithms \footnote{Roughly speaking, we use conventional algorithms to mean explicitly specified procedures that can be implemented in conventional software without relying on a learned model.} have been studied in toy transformer models \citep{nanda2023progressmeasuresgrokkingmechanistic, zhong2023clockpizzastoriesmechanistic}, there has been limited study into how conventional algorithms are used (or not) in LRMs and what role the CoT actually plays in producing the answer. Therefore, we propose the following research question (RQ) for LRMs:
\begin{quote}
    \textbf{RQ}: How does post-training change the decodability of latent algorithmic state and the functional role of verbalized algorithmic state in the chain of thought?
\end{quote}

By algorithmic state, we mean known intermediate variables in an algorithm, such as carries in addition. Latent algorithmic state refers to their internal representation in a language model, potentially associated with circuits \citep{elhage2021mathematical} or steering directions \citep{turner2024steeringlanguagemodelsactivation}, while verbalized algorithmic state refers to their explicit expression in the CoT. We focus on this distinction because it allows us to contrast different means of computation in a language model. Moreover, we can test the causal role of these representations by intervening on them and measuring whether subsequent computation changes as predicted by the corresponding algorithm. We do not, however, assume that the model implements a unique conventional algorithm, nor do we eliminate the possibility of other mechanisms, such as heuristics, working alongside it.

In order to study the RQ, we use simple addition problems of 2, 4, and 6 digits. Addition is desirable as a problem set here because of several properties: (1) conventional algorithmic state is simple and always known (per digit results and carries); (2) the algorithmic state is local and causal; (3) No deep knowledge is required nor can simple tricks be used to solve it. Further, addition may be used in downstream tasks where there is no correct answer but an exact justification is needed, such as vaccine distribution, where arithmetic on population counts would likely be used.

To that end, we propose two sub-questions (SQs) to the main research question that focus on addition:

\begin{enumerate}
    \item \textbf{SQ1}: Can we identify internal representations corresponding to incoming and outgoing carry bits, and do interventions on those representations selectively affect the appropriate output digits and does identifiability change across post-training?
    \item \textbf{SQ2}: What functional role do reasoning tokens play (w.r.t addition) and how do these roles change across post-training?
\end{enumerate}

In order to study the SQs across post-training, we use the Olmo-3 model \citep{olmo2026olmo3}, as it is the only open-weight model currently available that has a reasoning model variant and the full provenance and checkpoint history of its post-training phases. To that end, we make the following contributions across successive post-training checkpoints:
\begin{enumerate}
    \item We perform a controlled longitudinal study of addition problems.
    \item We identify decodable carry information and test the limits of several interventions on its causal use.
    \item We identify semantically meaningful spans in the CoT and measure their influence on later reasoning and the final answer.
\end{enumerate}

\begin{figure*}[t]
    \centering
    \begin{subfigure}[t]{0.55\textwidth}
        \centering
        \includegraphics[width=\linewidth]{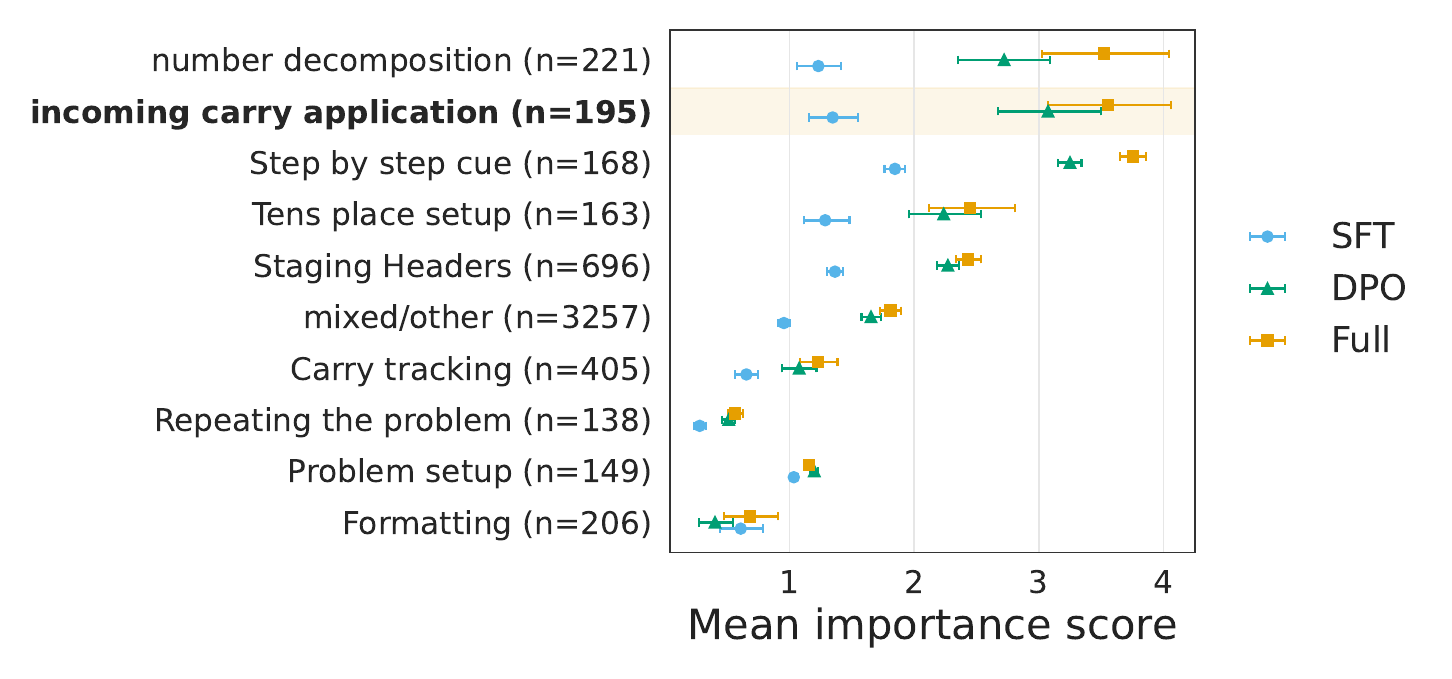}
        \caption{Semantic Importance}
        \label{fig:semantic-importance}
    \end{subfigure}
    \begin{subfigure}[t]{0.4\textwidth}
        \centering
        \includegraphics[width=\linewidth]{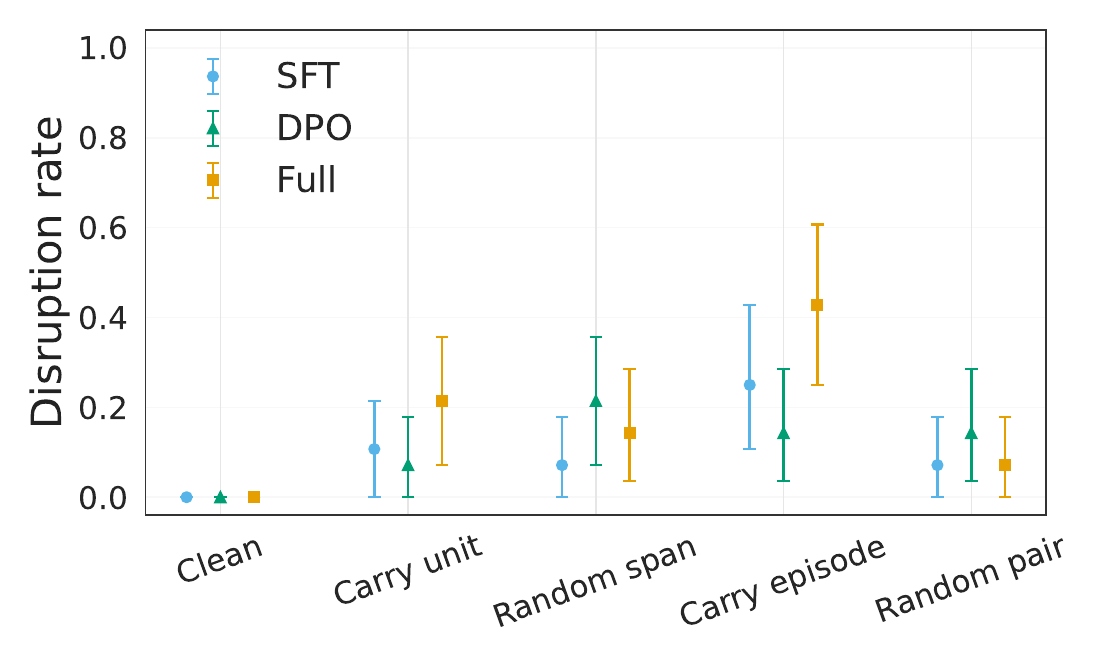}
        \caption{On policy carry-related disruption effect.}
        \label{fig:on-policy-disruption}
    \end{subfigure}

    \caption{
    Analysis of important spans in the CoT.
    \textbf{(a)} For each span in the CoT, we cluster based on semantic similarity and show the mean importance score derived from the \textit{Thought Anchors} method \citep{bogdan2025thoughtanchorsllmreasoning}. For a given span, we suppress all attention to that span and calculate its importance as described in \Cref{sec:semantic-spans}.
    \textbf{(b)} Each receiver continues from the same Full-generated prefix. We measure disruption as the percentage of correct cases where the digit in the column corresponding to the reasoning span changes. A carry unit is a reasoning span where a carry gets used directly; a carry episode is two spans, one where a carry is computed and another, the follow-up span, where that carry gets used. We include random spans and random pairs (episodes) as controls. Error bars are 95\% problem-level bootstrap confidence intervals for (a) and (b).
    }
    \label{fig:reasoning-spans-semantic-importance}
\end{figure*}

\Cref{fig:reasoning-spans-semantic-importance} previews our analysis of verbalized reasoning, showing how sensitivity to semantically meaningful CoT spans differs across checkpoints and how suppressing carry-related episodes affects subsequent answers. We describe the interventions and discuss these results in \Cref{sec:semantic-spans}.

\section{Related Work}

\paragraph{Algorithms in transformers.} Prior work has studied the emergence of conventional algorithms in toy transformers, particularly under \textit{grokking}. \citet{nanda2023progressmeasuresgrokkingmechanistic} identify a Fourier-based algorithm for modular arithmetic, while \citet{zhong2023clockpizzastoriesmechanistic} show that multiple distinct algorithms can solve the same task. Related work finds task-specific specialization in attention heads \citep{urdshals2025structuredevelopmentlistsortingtransformers,wang2024differentiationspecializationattentionheads} Evidence for comparable conventional algorithms in LLMs is more limited: \citet{lindsey2025biology} provide suggestive mechanistic results, but not definitive evidence. To our knowledge, no prior work has established a conventional algorithm in an LRM.

\paragraph{Addition and Arithmetic in LLMs.} 
There is sizeable literature showing that arithmetic-related information can be localized, decoded, ablated, or even used to steer individual output digits in ordinary pretrained LLMs \citep{nikankin2025arithmeticalgorithmslanguagemodels, deng2026languagemodelssymboliclearners, baeumel2025lookaheadlimitationmultioperandaddition, sun2025probingarithmeticerrorslanguage, du-etal-2025-fine, stolfo2023mechanisticinterpretationarithmeticreasoning}. Separately, there is strong evidence that carries are a difficult, non-local dependency even for frontier/reasoning models \citep{bie2026airithmetic}. Despite this, these works share a \textit{crucial limitation}: none have discovered a representation, circuit or otherwise, that modifies output digits correctly while correctly updating the resulting carry and preserving the rest of the computation; in fact, most of them do not consider problems with carries involved at all. As the carry is an essential part of the algorithm to solve addition, it calls into question whether the circuits discovered for addition so far are unique and faithful to what the model is really doing, or merely an incidental component of computation. Indeed, \citet{baeumel2025modulararithmeticlanguagemodels} show how patching their discovered mechanisms failed to appropriately adjust carries.
However, \citet{yan2025additionmovementsmappinglayerwise} do show that carry information is detectable via probes, but do not demonstrate any causal use of that information. 

\paragraph{The effectiveness of reasoning and RLVR.} RLVR based post training paved the way for the step changed in general performance shown in LRMs \citep{Guo_2025_deepseek_r1}. Despite this shift, work by \citet{venhoff2025basemodelsknowreason} shows that base models often contain the mechanisms needed for reasoning, and, alongside \citet{zhao2025activationcontrolefficientlyeliciting}, they show that we can activate such reasoning in the base models. While it appears clear that reasoning is important, it is less clear on how much of that reasoning is needed. Some work has shown that many reasoning tokens are unnecessary for producing the final answer, although this is less true for more difficult problems \citep{boppana2026reasoningtheaterdisentanglingmodel, scalena2026commitmentboundaryprobingepiphenomenal}.

Consequently, there is no consensus on whether models can do abstract, high level reasoning or are more reliant on memorization and heuristics, with proponents on both sides of the debate; likely the truth is a mix of both \citep{daliberti2026illusioninsightreasoningmodels, yang2025largelanguagemodelsperform, beckmann2026mechanisticindicatorsunderstandinglarge}.

\paragraph{CoT Faithfulness.} 
Often models will not faithfully describe their decision making or problem solving within their CoT. For example, a model may give a different answer when given a hint, but fail to mention that hint in its CoT \citep{turpin2023languagemodelsdontsay}. Likewise, this also applies to LRMs \citep{chen2025reasoningmodelsdontsay}. In this work, we consider elements of the CoT \textit{more likely} to be faithful when they are causally important to the final answer, but we do not address the problem of CoT faithfulness head on.

\section{Experimental Setup}
\begin{figure*}[t]
    \centering
    \begin{subfigure}[t]{0.3\textwidth}
        \centering
        \includegraphics[width=\linewidth]{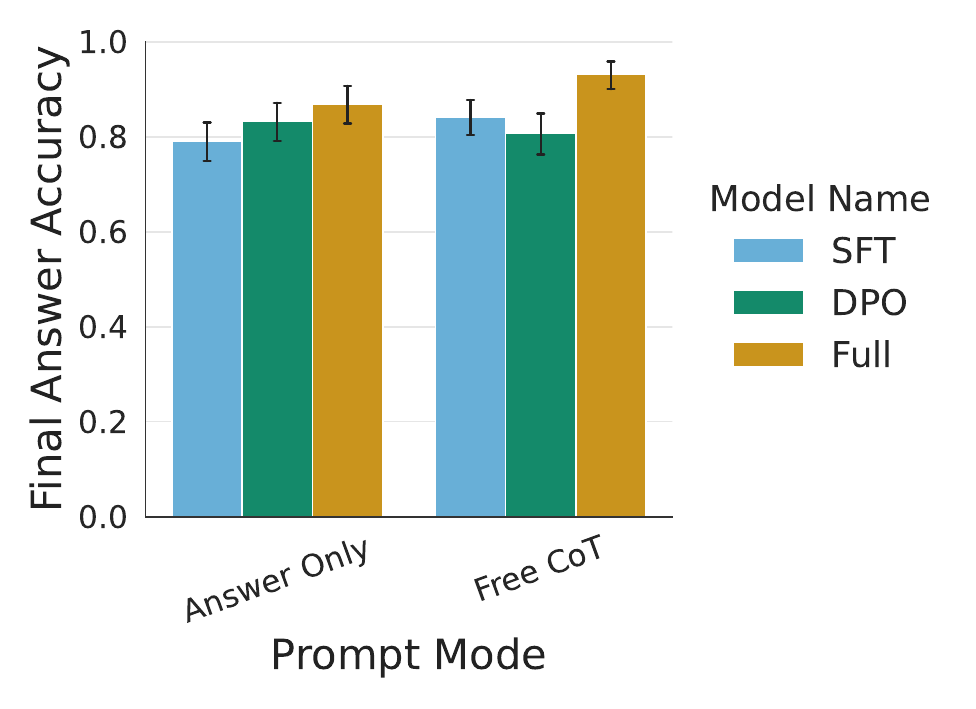}
        \caption{Prompt mode comparison}
        \label{fig:acc-comparison}
    \end{subfigure}
    \hfill
    \begin{subfigure}[t]{0.34\textwidth}
        \centering
        \includegraphics[width=\linewidth]{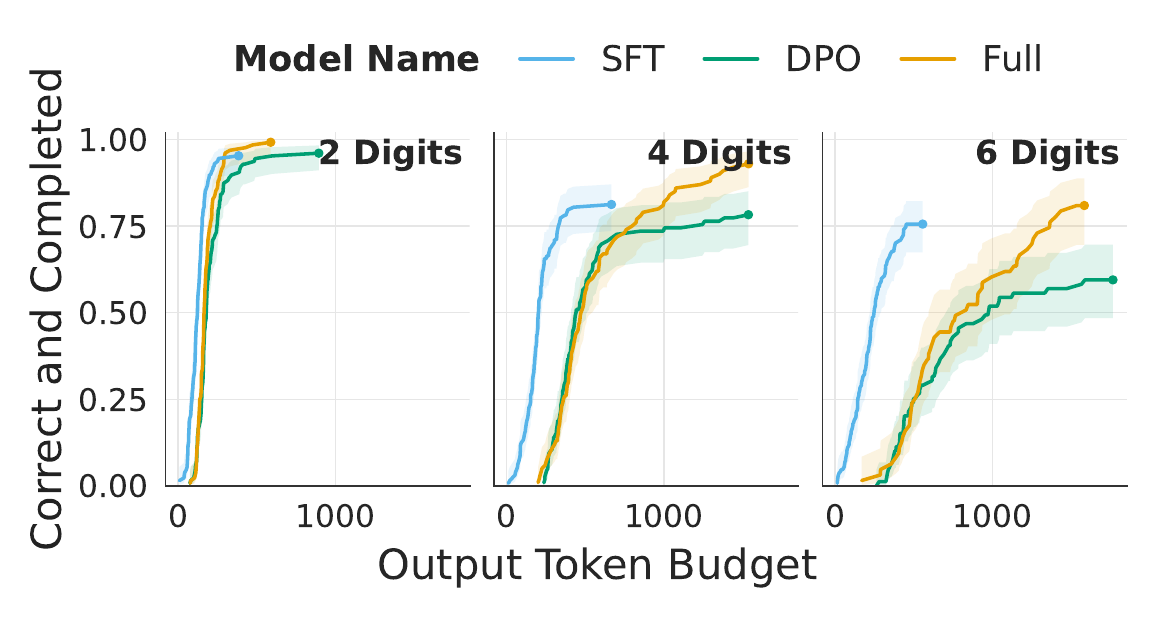}
        \caption{Token budget curves}
        \label{fig:token-budget-curves}
    \end{subfigure}
    \hfill
    \begin{subfigure}[t]{0.28\textwidth}
        \centering
        \includegraphics[width=\linewidth]{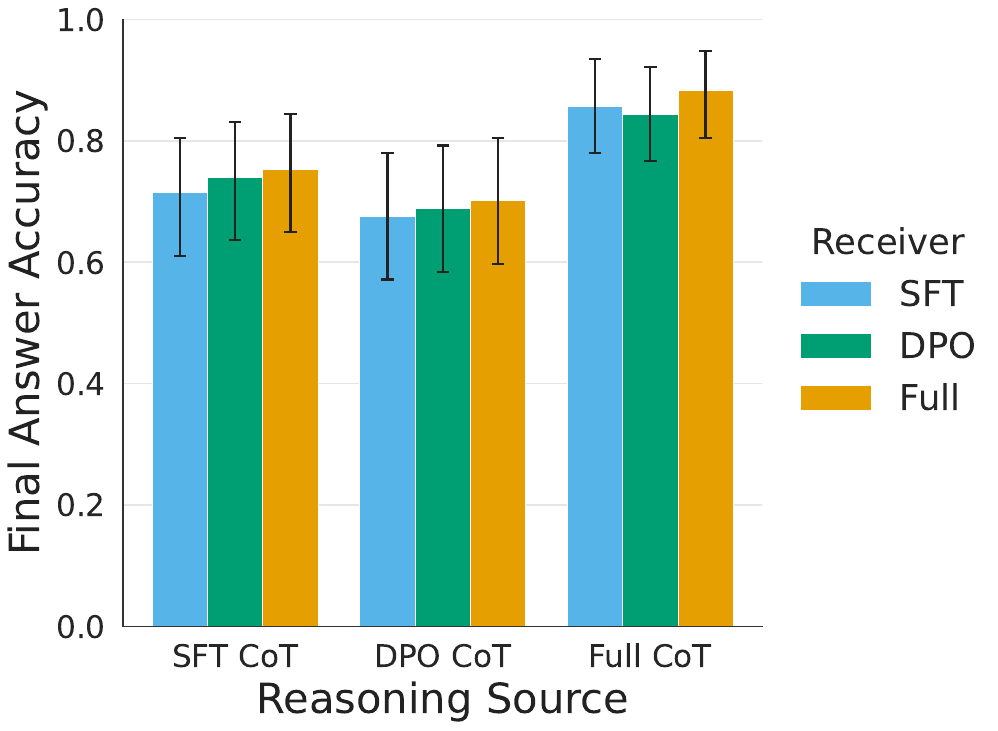}
        \caption{CoT crossover}
        \label{fig:cot-crossover}
    \end{subfigure}

    \caption{
    Analysis of chain-of-thought behavior across models. 
    \textbf{(a)} Final-answer accuracy for problems where both Answer Only and Free CoT terminate within 2048 tokens, paired separately for each checkpoint. Full gains 6.2 percentage points over its Answer Only baseline (adjusted $p=.034$); see \Cref{app:paired-prompt-modes} for sample counts and statistical testing details.
    \textbf{(b)} For each of \{2, 4, 6\} digits, we show the percentage of correct and completed samples with respect to the number of output tokens used, up to a maximum of 2048. Notably, the Full model saturates less with additional CoT tokens than the other two models
    \textbf{(c)} For each model, we produce an on-policy CoT and feed it to each of the other models using teacher forcing before asking for a final answer. All three models perform similarly when receiving a given CoT, but CoTs sourced from Full provide an advantage across all three models.
    Error bars are 95\% problem-level bootstrap confidence intervals for (a) and (c); the shaded region for (b) is the 95\% Wilson binomial confidence interval.
    }
    \label{fig:cot-combined-analysis}
\end{figure*}

\paragraph{Model choice.} In order to study the progression of post-training phases, we use the Olmo-3 family of models. We use the 7B parameter size variant and the \textit{Instruct} version of the model. Specifically, we focus on \texttt{Olmo-3-7B-Instruct-SFT}, \texttt{Olmo-3-7B-Instruct-DPO}, and \texttt{Olmo-3-7B-Instruct}. We refer to the models with shorthands SFT, DPO, and Full respectively. We use the \textit{Instruct} variant because it was trained to limit its verbosity \citep{olmo2026olmo3}; we found that output token usage for the \textit{Think} variant grew super-linearly with the number of digits in the problem, producing untenable outputs. While the type of "thoughts" produced in the \textit{Instruct} chain-of-thought may differ from the hidden \textit{Think} chain-of-thought (e.g. tokens enclosed in <think> tags), we leave the extent to which their causal roles are similar (or not) as an open question.

Notably, the progression of post-training in Olmo-3 proceeds as an SFT stage, a DPO stage (particularly on high quality preferences from a judge model), and a RLVR stage. The full contents of what data are used in each stage is described in the original paper, alongside other training details \citep{olmo2026olmo3}. We do note that a significant amount of math and reasoning data are included at each stage. In \Cref{app:memorization}, we find little exact overlap with the public post-training data for 4- and 6-digit evaluation equations, although this does not rule out pretraining exposure or learned arithmetic patterns.

\paragraph{Dataset Composition.} Our datasets composed of 2, 4, and 6 digit addition problems, balanced across the number of carries needed to solve the problem. For each problem, we use two different prompts to gather two answers: \textit{Answer Only}, where the model is asked for only the answer; and \textit{Free CoT}, where the model is asked to verbalize its reasoning step-by-step before providing a final answer. The prompt templates for both are available in \Cref{app:prompts}. Each analysis below typically uses a subset of these addition problems and the relevant data is described for each. For example, some analysis is limited to 4 digits as the Full model often fails to terminate its output for 6 digit problems (that is, its output would degenerate and fail to make progress). We also provide results for ablations on the problem presentation and formatting, including tokenization, in \Cref{app:additional-results}. A complete dataset description is available in \Cref{app:dataset-description}.

\paragraph{Scope of the analyses.} Our focus is on computation within completed reasoning traces; we treat degeneration and failure to terminate as a separate aspect of generation behaviour. We note that the Full model would sometimes fail to stop reasoning even with very large token budgets; still, we consider the naturally terminating cases to be the most accurate examples of nominal behavior and therefore focus our analysis on them.

 \paragraph{Definition of algorithmic state for addition.} In terms of algorithmic state, we define it as a state representing any computationally useful variables in a conventional algorithm. We do not restrict \textit{what form} the representation takes. For addition, there are two computationally useful variables we identify: the per-digit outputs and the per-digit carry. We identify a representation with an algorithmic state through both correlation and causal use that corresponds to the same effect on the computational variables. That is, the representation should ideally demonstrate the ability to causally mediate the relevant computational variables.

\section{Post-Training Changes CoT Usefulness}

To establish a baseline for both \textbf{SQ1} and \textbf{SQ2}, we compare final-answer accuracy with and without CoT on problems where both prompt modes produce completed responses. \Cref{fig:acc-comparison} shows these results, with 383, 312, and 291 retained problems for SFT, DPO, and Full respectively. Only the Full model has a statistically significant gain in accuracy using the CoT, but the \textit{difference} in gain between Full and SFT is not statistically significant. Statistical details are provided in \Cref{app:paired-prompt-modes}.

Interestingly, \Cref{fig:token-budget-curves} shows that Full, and to some extent DPO, continue to produce additional correct and completed answers as the output-token budget increases, particularly for more difficult problems (i.e. more digits). Thus, a longer CoT is necessary to realize some of its successful reasoning trajectories. This suggests \textit{a} functional role for longer generations, although the curves alone do not establish that longer generations are always necessary nor \textit{which} reasoning steps are essential.

Finally, we introduce CoT crossover in \Cref{fig:cot-crossover}, where each model receives, via teacher forcing before the final answer, the CoT produced by another model, in order to estimate the effectiveness of \textit{producing} the CoT as compared to \textit{using} it. We find that for the \textit{same source}, all receivers show similar accuracy. However, for CoTs from the Full model, all receivers obtain higher final-answer accuracy compared to the SFT or DPO sourced CoTs.

\section{Probes of Algorithmic State are Stable Across Post-Training}
To find whether a representation of algorithmic state is present in a models activation (as asked by \textbf{SQ1}), we use linear probes \citep{alain2018understandingintermediatelayersusing} of incoming carry, outgoing carry, and output digit across all digits in a problem. We perform probing at every layer throughout each model and at important tokens in the prompt and part way through the CoT. 

\begin{figure*}
    \centering
    \includegraphics[width=\textwidth]{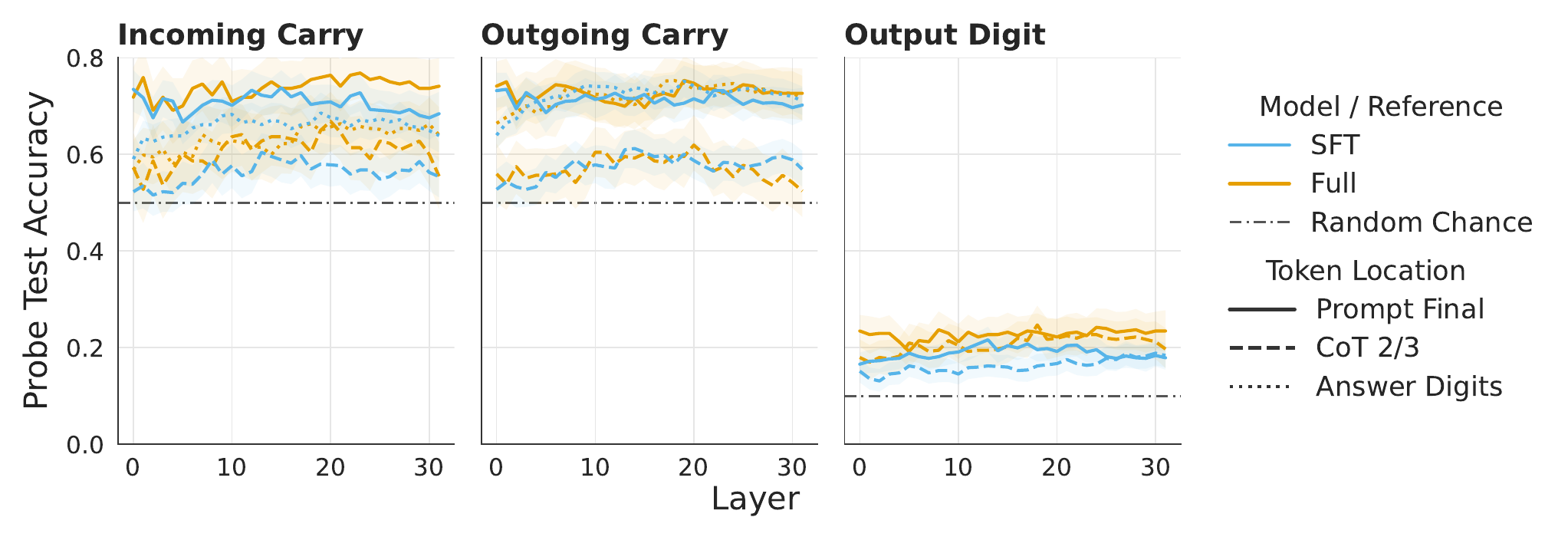}
    \caption{
    We measure probe test accuracy at three token locations: the final token of the prompt; 2/3rd of the way through the CoT; and at the answer digit token. We train probes separately for three variables: the incoming carry at a digit; the outgoing carry at a digit; and an output digit. No probes are shown for finding the output digits at the answer digit token, as the answer would often be already available (up to tokenization effects). Probes are trained separately for each variable, layer, digit, and location. The results shown are averaged across digits. Overall, the probe test accuracy remains stable across layers and very similar across models, while showing only minor differences at different token locations. Shaded regions show 95\% confidence intervals from problem-level bootstrap resampling. Dash-dot lines show uniform random-chance accuracy.}
    \label{fig:probe-profile}
\end{figure*}

We essentially find that there is little difference in probe accuracy across both layers and tokens in the models tested \footnote{We only test SFT and Full due to the expensive permutations of layers and tokens and due to the highly similar probe results between these models.}, particularly for incoming and outgoing carry, which remain stable across layers and tokens. However, we do note that there is small decrease in probe test accuracy at the CoT location. The results are shown in \Cref{fig:probe-profile}. More detailed results and training procedure for the probes are described in \Cref{app:probes}.

We also test using probe directions as steering vectors to manipulate per digit carries in a predictable way. Doing this, we find a negative result: the probe-implied steering directions are not able to causally manipulate the per digit carry correctly. Details are provided in \Cref{app:probe-steering}.

\section{Internal Interventions Do Not Reliably Control Carry State}
Following the negative results from probe-based steering, we tested two further ways of manipulating internal carry representations to better answer \textbf{SQ2}. Boundless Distributed Alignment Search (DAS) \citep{wu2024interpretabilityscaleidentifyingcausal} learns which part of an activation to swap between problems to produce a chosen change in the answer. Activation patching \citep{heimersheim2024useinterpretactivationpatching} directly replaces activations at selected layers and token positions with those from another problem. In both cases, we used matched problem pairs with opposite incoming carries at a chosen column.

A successful intervention would make the model continue as though the incoming carry had changed from 0 to 1, or vice versa. We chose problems where this change should affect only the answer digit in that column, leaving the outgoing carry and all other answer digits unchanged. Thus, simply changing the output would not count as success: the answer needed to change in the specific way predicted by the carry swap.

Neither method achieved this reliably. On 64 held-out controlled pairs, DAS produced the predicted complete answer in 1.6\% of cases for SFT and 12.5\% for Full, and did not consistently preserve the other digits (see \Cref{app:das}).

For activation patching, we first tested different layers and lengths of text on a development set of controlled problems. Our primary configuration replaced the activations for the eight tokens immediately before the target-column calculation at layer 27. On 64 separate controlled pairs, this intervention never produced the complete answer expected from changing the carry (see \Cref{app:activation-patching}).


Ultimately, carry information is decodable (like \citet{yan2025additionmovementsmappinglayerwise}), but our tested directions and patches do not provide reliable causal control (mirroring difficulties reported by \citet{baeumel2025modulararithmeticlanguagemodels}). Distributed computation or non-linear representations are possible explanations, consistent with \citet{wurgaft2026manifoldsteeringrevealsshared}, but these null results do not distinguish them from limitations of the intervention locations, optimization, or assumed carry abstraction. Details of our causal intervention experiments can be found in \Cref{app:causal-interventions}.

\section{Post-training Changes the Causal Importance of Semantically Meaningful Spans in the CoT}\label{sec:semantic-spans}

\paragraph{Attention suppression and importance.} To address \textbf{SQ2}, we study the importance of different spans in the CoT by following \citet{bogdan2025thoughtanchorsllmreasoning}. Specifically, we apply teacher forcing of a naturally produced CoT while suppressing attention across all layers to a particular span, and measuring the downstream change in probabilities.

We define spans as sections of the CoT that roughly correspond to one semantic 'unit'. We make this distinction as compared to using, say, sentences, because models often use markdown and structure the output with headings, use bullet points, and short phrases. We therefore separate out each heading, bullet point, and sentence (within a paragraph) into a distinct unit.

More specifically, we define importance as 

\begin{align}
    \mathrm{KL}_{\mathrm{unit}}(i) &= \max_{j > i}\mathrm{meanTokenKL}(i \to j), \\
    \mathrm{KL}_{\mathrm{ans}}(i) &= \mathrm{meanAnswerKL}(i), \\
    \mathrm{importance}(i) &= \max\left( \mathrm{KL}_{\mathrm{unit}}(i), \mathrm{KL}_{\mathrm{ans}}(i) \right).
\end{align}

Each token-level divergence is
\(D_{\mathrm{KL}}(P_{\mathrm{clean}} \,\|\, P_{\mathrm{suppressed}})\). The first term (meanTokenKL) is the largest mean per-token KL over any single later
reasoning unit; the second (meanAnswerKL) is the mean per-token KL over the teacher-forced factual-answer tokens. The score is measured in nats and identifies units whose suppression strongly changes at least one later reasoning unit or the answer. We take the maximum because a span may strongly affect a later reasoning step without strongly affecting the final answer, or vice versa. A large score means that suppressing attention to the span substantially changes the model’s predictions for at least one later reasoning step or the answer; it does not necessarily mean the answer becomes less accurate.

\paragraph{Semantic clustering.} Further, we perform semantic clustering of reasoning spans using TF–IDF unigrams and bigrams for K-Means clustering. After clusters were generated this way, we performed manual audits on the clusters, merging some, and assigning semantic labels to all the remaining clusters. We also singled out some specific categories for clustering, like punctuation/formatting and, as it is of special interest, the use of carries in a span. For more detailed explanations of labels, examples from each cluster, and how clusters were chosen, see \Cref{app:clustering-addition}.

\paragraph{Results from suppression and clustering.} With these tools, we measured the mean importance score for naturally generated CoTs and performed semantic clustering, as shown in \Cref{fig:semantic-importance}. Notably, we find that algorithmically meaningful clusters, like incoming carry application and number decomposition, receive higher importance scores in DPO and Full than in SFT. We also see that spans like problem repetition and formatting receive low scores for all three models. This result is suggestive of the idea that the DPO and Full models are more dependent on spans which contain algorithmically useful content.

\paragraph{On-Policy re-generation.} In order to assess the effects on newly generated continuations, we again applied attention suppression, but allowed each receiver to generate from the same Full-produced prefix while continuing the suppression throughout generation. Our metric of interest here is the disruption rate, which we define as the percentage of problems were the digit in column being referenced in a reasoning span changes (we limit to problems where the final answer was correct under clean/natural generation). 

As shown in \Cref{fig:on-policy-disruption}, we find that, actually, suppressing the application of a carry itself does not result in substantially more disruption than suppressing a random span. 

We then hypothesized that the model could likely recover or otherwise infer the information from the suppressed span by attending to the span directly before the application of the carry. In other words, if the carry span was "The hundreds place is 2 + 4 + 1 (carry)" then, often, the span before would be something to the effect of "The output for the tens place is 12, so the output is 2 with a carry of 1 for the hundreds place". As such, we decided to also suppress attention to the span directly before the application of a carry, which we refer to as a "carry episode." As a control for the episode, we also test random pairs of spans. 

For these episodes, we find something different: the Full model shows significantly more disruption when the carry episode is suppressed when compared to a random pair control. SFT also shows increased disruption, but the effect falls within uncertainty. 

For Full, the importance scores and generation results tell a consistent story: carry-related spans strongly affect predictions along a fixed CoT, and suppressing a carry episode disrupts the model's continuation more than suppressing a random pair of spans. 

For SFT, however, lower importance scores coexist with similar observed levels of disruption during generation. One possible explanation is that changes which appear small when the original CoT is held fixed can build up as the model generates its own continuation. The interventions also differ: the importance scores measure suppression of individual spans, whereas episode suppression removes access to both the carry-producing and carry-applying spans. This may limit opportunities to recover the carry from nearby text. Lower importance scores therefore need not imply that a model can continue successfully without the corresponding information. 

These interpretations remain tentative: the episode effect relative to random controls is uncertain for SFT, and we do not establish a significant difference in that effect between SFT and Full. Episode suppression also disrupts other digits and termination (\Cref{fig:on-policy-secondary-outcomes}), so the effects cannot be attributed solely to the carry value.

\paragraph{Extending to subtraction.} To assess how the results of semantic importance generalize to other algorithmic tasks, we extend the attention suppression and clustering experiment to subtraction. Excepting the change of problem type and changing carries to borrows, the setup remains the same. Details on clustering, including examples and explanation of semantic labels, are provided in \Cref{app:clustering-subtraction}.

\begin{figure}
    \centering
    \includegraphics[width=\linewidth]{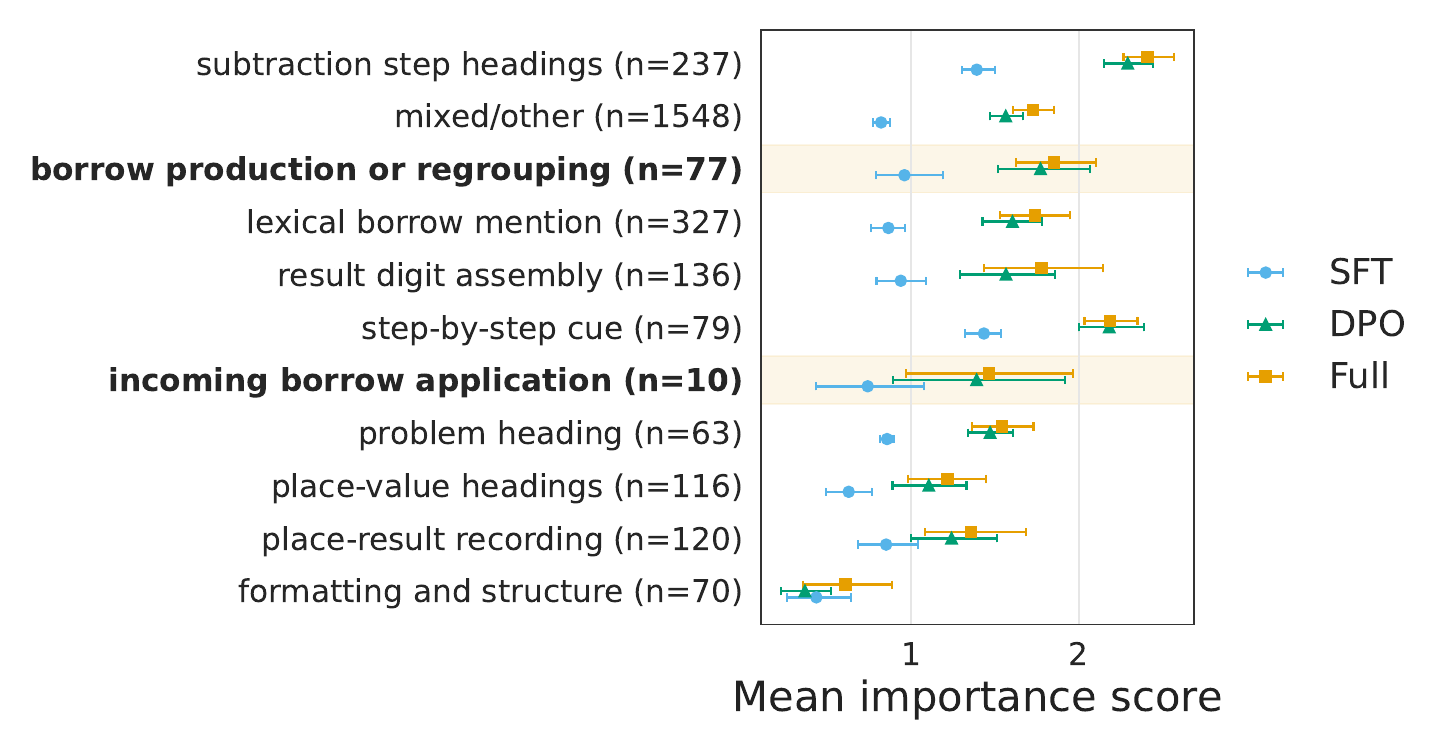}
    \caption{Importance scores for semantic clustering of reasoning spans on subtraction problems. The top cluster by importance, subtraction step headings, take the form of "Step <num>: Subtract the <units/tens/hundreds/...> place". Specified clusters involving the use of borrows are highlighted; some other mentions of borrow fall into the lexical borrow mention cluster. More details on clustering are provided in \Cref{app:clustering-subtraction}. Error bars are 95\% problem-level bootstrap confidence intervals.}
    \label{fig:semantic-importance-subtraction}
\end{figure}

We find, as shown in \Cref{fig:semantic-importance-subtraction}, that the importance of borrow related spans is not as strongly different across the models as it is for carries in addition, although the same pattern of DPO and Full having higher importance scores across the board remains. Other work, in particular \citet{jobanputra2025llmssubtractnumbers}, show that models do not perform subtraction as well as they perform addition. This suggests that the dependence on algorithmically meaningful spans in the CoT is task dependent, and that different algorithms are treated with differing levels of importance. Given the uneven nature of task ability in LLMs and LRMs, we are not surprised to find that different algorithmic tasks admit varying levels of reliance on algorithmic state in the CoT across models \citep{dellacqua2023navigating}.

\section{Discussion}
Our results suggest that post-training changes how intermediate information is made available and used during reasoning. Carry information is similarly decodable in SFT and Full, while Full produces traces that are more useful across receiving checkpoints. One interpretation is that post-training improves how the model supplies context for its own later computation, drawing on capabilities already available at SFT \citep{venhoff2025basemodelsknowreason}. Written intermediate results could reduce the need to recover or recompute information internally. This makes latent and verbalized state complementary parts of a reasoning process: changing what is written can change what subsequent internal computation needs to accomplish. The crossover results support the usefulness of that context, while leaving open whether receivers calculate from it, check against it, or extract an answer.

The suppression experiments suggest that this context may support computation across several reasoning steps. For Full, removing a carry episode has a stronger effect than matched random suppression, even though removing the carry application alone has little additional effect. This is consistent with information remaining available elsewhere in the trace. A statement can therefore be individually dispensable while contributing to a useful reasoning process, complicating interpretations of local intervention failures as evidence of performative reasoning \citep{boppana2026reasoningtheaterdisentanglingmodel,scalena2026commitmentboundaryprobingepiphenomenal}. The broader disruption caused by episode suppression prevents us from attributing the effect solely to carry recovery, but motivates studying how information is maintained across a sequence of steps.

Dependence on this context does not guarantee that it is used well. DPO combines relatively high importance scores with lower conditional accuracy, while SFT's lower scores coexist with similar observed disruption during generation. These patterns separate influence on predictions from successful continuation. Written reasoning may help preserve a correct intermediate result, but it can also preserve a mistake; small changes along a fixed trace may become consequential when the model generates subsequent steps itself. Post-training could therefore improve reasoning by changing when intermediate information is trusted, checked, or revised. The residual and J-Lens comparisons identify locations where processing differs, but neither these readouts nor our unsuccessful carry interventions establish which of these functions changes.

This perspective also gives reasoning length a role beyond the number of distinct computational steps. Additional budget is needed for some successful Full trajectories under its Free CoT policy, and part of that budget could support repetition, checking, or recovery. Training with stronger pressure on length might remove these opportunities even when clean accuracy is preserved. A useful test would therefore ask whether shortened traces remain robust to missing or incorrect intermediate information. The weaker carry-to-borrow correspondence in subtraction further suggests that the value of such strategies depends on the computation being performed. Understanding post-training may therefore require examining how it balances progress, verification, and recovery across tasks.

\section{Conclusion}

In this work, we study how post-training changes carry decodability and the functional role of verbalized reasoning in language models. Using progressive checkpoints of Olmo-3 and addition as a setting where the relevant algorithmic state is precisely known, we distinguish decodable information from demonstrated causal control. Carry information is similarly decodable in SFT and Full, yet the tested probe directions, learned subspaces, and activation patches do not reliably manipulate carries while preserving the rest of the answer.

At the same time, the role of the CoT differs across checkpoints. In the completed-response comparison, Full gains accuracy from CoT, and its traces yield higher accuracy than SFT- or DPO-sourced traces across all receivers in the crossover experiment. Suppressing carry episodes also disrupts Full more than matched random controls. However, higher suppression importance need not coincide with higher accuracy, as shown by DPO, and the subtraction results suggest that these patterns are task dependent. Accuracy alone therefore does not tell us how strongly a model depends on its written reasoning.

Taken together, these findings motivate studying the production, consumption, and causal influence of written reasoning separately. Our results are compatible with the idea that post-training changes how existing computational resources are recruited, although we do not identify that mechanism. We further speculate that Chain-of-thought is therefore neither straightforwardly necessary as a means of computation nor merely performative: under some training regimes and for some problems, written reasoning becomes causally important as part of a broader computational process. Establishing precisely how carry-related context mediates conventional arithmetic, and how this interacts with latent computation, remains an important next step.

\section*{Limitations}
We limit our study to one model family due to its unique nature and we primarily study addition. Other model families and algorithmic problems may show different results. Likewise, important training details such as number of training steps, optimization details, and hyper-parameter choices may affect the results, as well as the data mixture. The analysis presented here is therefore mostly post-hoc with regards to training: differences between successive checkpoints do not isolate the training objective from the data and other training choices. The completion restrictions also limit the population to which our reasoning analyses apply. Separately, our negative results on finding causally valid interventions for carries in addition do not rule out that such interventions exist or that the methods we used could find them under other settings. We also note that model size is not studied here and likely plays an interesting role when considering algorithmic computation; we leave that to future work.


\section*{Ethics Statement}
This research intends to understand the mechanisms of algorithmic computation in language models. We do not make claims about strongly controlling said computation or adversarially manipulating it, and therefore we do not associate risks with this research. Future work, however, which gives users control over reasoning, may pose dual-use risks.

\section*{Usage of AI}
AI was used to generate code for experiments, provide feedback on writing, and for collating data for figures and tables. AI was also used for planning and orchestrating compute usage.

\bibliography{references_updated}

\appendix

\begin{table*}[t]
    \centering
    \small
    \setlength{\tabcolsep}{4pt}
    \renewcommand{\arraystretch}{1.08}
    \begin{tabularx}{\textwidth}{@{}p{0.18\textwidth}p{0.49\textwidth}X@{}}
        \toprule
        Dataset & Construction & Size used in the paper \\
        \midrule
        Behavioural addition
        & Two-, four-, and six-digit addition. The standard set contains four controlled carry structures and a carry-balanced random sample. Separate random samples test delimited operands and reversed answer digits.
        & 384 standard problems and 768 standard prompt instances; 1,536 prompt instances including the two format conditions. \\

        Probe addition
        & Two-, four-, and six-digit addition, balanced by carry count and divided into disjoint training and test sets. Every problem appears under both prompt conditions.
        & 384 problems: 192 training and 192 test; 768 prompt instances in total. \\

        Paired Free CoT addition
        & Four- and six-digit problems for which natural reasoning traces are collected from SFT, DPO, and Full. Full is used to choose the initial set; analyses comparing reasoning sources use only problems with a usable trace from every checkpoint.
        & 384 candidates; 124 problems in the initial set; 77 problems in the three-source subset. \\

        On-policy addition
        & Fresh four-digit Full traces containing an identifiable use of an incoming carry, an earlier statement that establishes the carry, sufficient following context, and matched control spans.
        & 192 candidates; 51 problems satisfy all analysis requirements. \\

        Paired Free CoT subtraction
        & Four-digit subtraction, balanced across zero, one, two, and three borrows. Full and SFT provide paired reasoning traces; all three checkpoints are evaluated as receivers.
        & 192 candidates; 55 initially selected problems; 47 problems with usable traces from both sources. \\
        \bottomrule
    \end{tabularx}
    \caption{Datasets used in the paper. We distinguish different counts of arithmetic problems, prompt instances, and generated reasoning traces.}
    \label{tab:dataset-collections}
\end{table*}

\section{Complete Dataset Description}
\label{app:dataset-description}

All arithmetic examples are generated in base ten. Each example records the two operands, the exact result, and the state of the column-wise algorithm. For addition, this state includes the incoming and outgoing carry at each column, the positions and total number of carries, and the longest sequence of
consecutive carries. The subtraction data record the analogous borrow variables. Digit length always refers to the width of each operand, not the
width of the result. Problems are sampled before model inference, and a failed or incomplete model response is not replaced with a new problem.

We use two prompt conditions throughout the paper. \textit{Answer Only} asks the model to return only the final result. \textit{Free CoT} asks the model to reason step by step before giving a final result. A \textit{prompt instance} is one arithmetic problem under one prompt and formatting condition. A \textit{reasoning trace} is one generated Free CoT response. The main datasets are summarized in \Cref{tab:dataset-collections}; subsequent paragraphs give the construction and selection details.

\paragraph{Behavioural addition.}
The standard set contains 128 problems at each of 2, 4, and 6 digits. Within
each digit length, there are 16 no-carry problems, 16 problems with one
isolated carry, 16 problems in which a carry propagates across every operand
digit, 16 carry-distractor problems, and 64 random problems. Carry distractors
contain locally carry-suggestive digit patterns without a long carry chain.
The random problems are balanced by the number of carry-producing columns.
Every standard problem is presented once under Answer Only and once under Free
CoT, giving 768 standard prompt instances per checkpoint.

The two format conditions use separate random samples. Each contains 64 problems at each digit length and both prompt conditions, for 384 prompt instances. In the delimited input condition, adjacent operand digits are separated by \texttt{|}; the answer remains conventionally formatted. In the least-significant-digit (LSD) first condition, the operands are conventional but the answer digits must be written from right to left. Together, the standard and format-specific sets contain 1,536 prompt instances. Because the arithmetic examples differ between conditions, the format results are unpaired robustness comparisons.

\paragraph{Probe addition.}
The probe dataset contains separate training and test sets, each with 64
problems at every digit length. Carry counts are balanced within each split,
and every problem appears under both prompt conditions. The probe results in
the main paper use the Free CoT instances. Before filtering for naturally completed samples, there are 192 training and 192 test examples per checkpoint; SFT retains all of them, while Full retains 125 training and 116 test examples (\Cref{tab:probe-sample-counts}). Activations are collected from SFT and Full.

Probes are fitted separately for each predicted variable, layer, arithmetic
column, and location; reported curves then average test accuracy
over arithmetic columns.

\paragraph{Paired natural reasoning traces.}
The source--receiver experiments require complete reasoning traces with
well-defined internal and final-answer boundaries. Here, the \textit{source}
is the checkpoint that writes the reasoning trace, and the \textit{receiver}
is the checkpoint evaluated after that trace is supplied by teacher forcing.
Trace eligibility requires natural termination, a parseable final answer, and
unambiguous reasoning boundaries.

Full is first run on 128 four-digit and 256 six-digit candidate problems with
a 2,048-token budget. The first eligible problems in the fixed candidate order
are retained, producing 64 four-digit and 60 six-digit problems. SFT and DPO
are then run on these same problems. The numbers of eligible four- and
six-digit traces are 61 and 60 for SFT, 47 and 32 for DPO, and 64 and 60 for
Full. Forty-five four-digit and 32 six-digit problems have an eligible trace
from all three checkpoints. This 77-problem subset is used whenever reasoning
sources are compared directly, giving 231 source traces. Analyses that do not
require all three sources use all 324 individually eligible traces.

\paragraph{Reasoning units and attention suppression.}
For the teacher-forced attention-suppression analysis, both the SFT and Full
trace for a problem must satisfy the same termination and boundary criteria
within the 1,024-token analysis window. This leaves 59 four-digit and 42
six-digit problems. Each trace is divided into semantically coherent reasoning
units: headings, bullet points, sentences, and short standalone phrases are
treated as separate units. The resulting 202 source traces contain 5,598
units, with 3,038 units from four-digit problems and 2,560 from six-digit
problems. SFT and Full provide the written traces, while SFT, DPO, and Full are
all evaluated as receivers of the same trace and suppression intervention.

The on-policy analysis uses a separate set of 192 four-digit problems. Full
generates a fresh trace for each problem. Fifty-one traces contain an explicit
incoming-carry application, an earlier unit that establishes the same carry,
sufficient later text for evaluation, and suitable random control units. The
same 51 problems are evaluated under clean generation, suppression of the
carry-application unit, suppression of a matched random unit, suppression of
the two-unit carry episode, and suppression of a matched random pair.

The residual-stream and Jacobian Lens (J-Lens) analysis uses a stratified subset of the
teacher-forced attention-suppression data. It contains 30 incoming-carry
applications from 23 problems: eight SFT/four-digit, eight SFT/six-digit,
eight Full/four-digit, and six Full/six-digit units. Each selected unit has
resolvable token positions immediately before the carry value and the resulting value.

\paragraph{Subtraction.}
The subtraction candidate set contains 192 four-digit problems, with 48
problems at each of zero, one, two, and three borrows. Full is used to select up
to 16 naturally completed traces in each borrow stratum under a 1,024-token
budget, after which SFT is run on the same problems. Fifty-five problems have
an eligible Full trace, and 47 have eligible traces from both SFT and Full. The
paired set contains 14, 14, 11, and 8 problems across the four borrow strata,
respectively. Its 94 source traces contain 2,783 reasoning units. As in the
addition analysis, SFT and Full provide the written traces and all three
checkpoints are evaluated as receivers.

\begin{table*}[t]
    \centering
    \footnotesize
    \setlength{\tabcolsep}{4pt}
    \setlength{\extrarowheight}{5pt}
    \begin{tabularx}{\textwidth}{@{}p{0.22\textwidth}X p{0.27\textwidth}@{}}
        \toprule
        Paper analysis & Dataset subset & Effective sample \\
        \midrule
        Behaviour and CoT usefulness (\Cref{fig:acc-comparison,fig:token-budget-curves})
        & Standard behavioural addition. Accuracy compares paired completed responses; token-budget curves use Free CoT.
        & 384 problems attempted per prompt condition and checkpoint. The paired accuracy comparison retains 383 SFT, 312 DPO, and 291 Full problems. \\

        CoT crossover (\Cref{fig:cot-crossover})
        & Problems with an eligible SFT, DPO, and Full trace. Every source trace is supplied to every receiver.
        & 77 problems: 45 four-digit and 32 six-digit; 77 observations in each source--receiver cell. \\

        Linear probes (\Cref{fig:probe-profile})
        & Free CoT instances from the disjoint probe training and test sets; SFT and Full only.
        & SFT: 192 training and 192 test examples. Full: 125 training and 116 test examples after the completion filter. \\

        Probe steering and activation interventions (\Cref{app:probe-steering,app:causal-interventions})
        & Probe steering reuses the probe dataset. Residual interventions use a separate shared Free CoT set. Distributed alignment search and activation patching use controlled carry pairs.
        & Residual interventions: 100 problems, of which 89 yield eligible cases. Alignment search: 1,024 training, 256 tuning, and 64 evaluation pairs. Activation patching: 16 development and 64 evaluation pairs. \\

        Answer emergence and commitment
        & Paired trajectories use the three-source subset; single-source trajectory summaries use every eligible trace.
        & 77 problems and 231 traces for paired comparisons; 324 traces for single-source summaries. \\

        Semantic importance in addition (\Cref{fig:semantic-importance})
        & Paired SFT/Full traces, divided into reasoning units, with SFT, DPO, and Full as receivers.
        & 101 problems, 202 traces, and 5,598 reasoning units. \\

        On-policy suppression (\Cref{fig:on-policy-disruption})
        & Fresh four-digit Full traces with an identifiable carry application, carry-producing context, and matched controls.
        & 51 analysed problems from 192 candidates; all conditions are evaluated with each receiver. \\

        Residual-stream and J-Lens comparison (\Cref{fig:residual-readout})
        & Stratified incoming-carry applications from the addition attention-suppression set, using SFT and Full source traces.
        & 30 reasoning units from 23 problems across two digit lengths and two source checkpoints. \\

        Semantic importance in subtraction (\Cref{fig:semantic-importance-subtraction})
        & Paired SFT/Full traces from the four borrow-count strata, with all three checkpoints as receivers.
        & 47 problems, 94 traces, and 2,783 reasoning units. \\

        Memorization and format robustness (\Cref{app:memorization,app:additional-results})
        & Standard Answer Only equations and their arithmetic-equivalent rewrites; separately sampled standard, delimited-input, and reversed-answer conditions.
        & 384 original equations and 1,084 valid rewrites; 1,536 behavioural prompt instances across formatting conditions. \\
        \bottomrule
    \end{tabularx}
    \caption{Dataset subset used by each analysis. Counts refer to arithmetic
    problems unless prompt instances, reasoning traces, reasoning units, or
    controlled intervention pairs are stated explicitly.}
    \label{tab:analysis-dataset-map}
\end{table*}

\section{Memorization}
\label{app:memorization}

\begin{figure}[h]
    \centering
    \includegraphics[width=\linewidth]{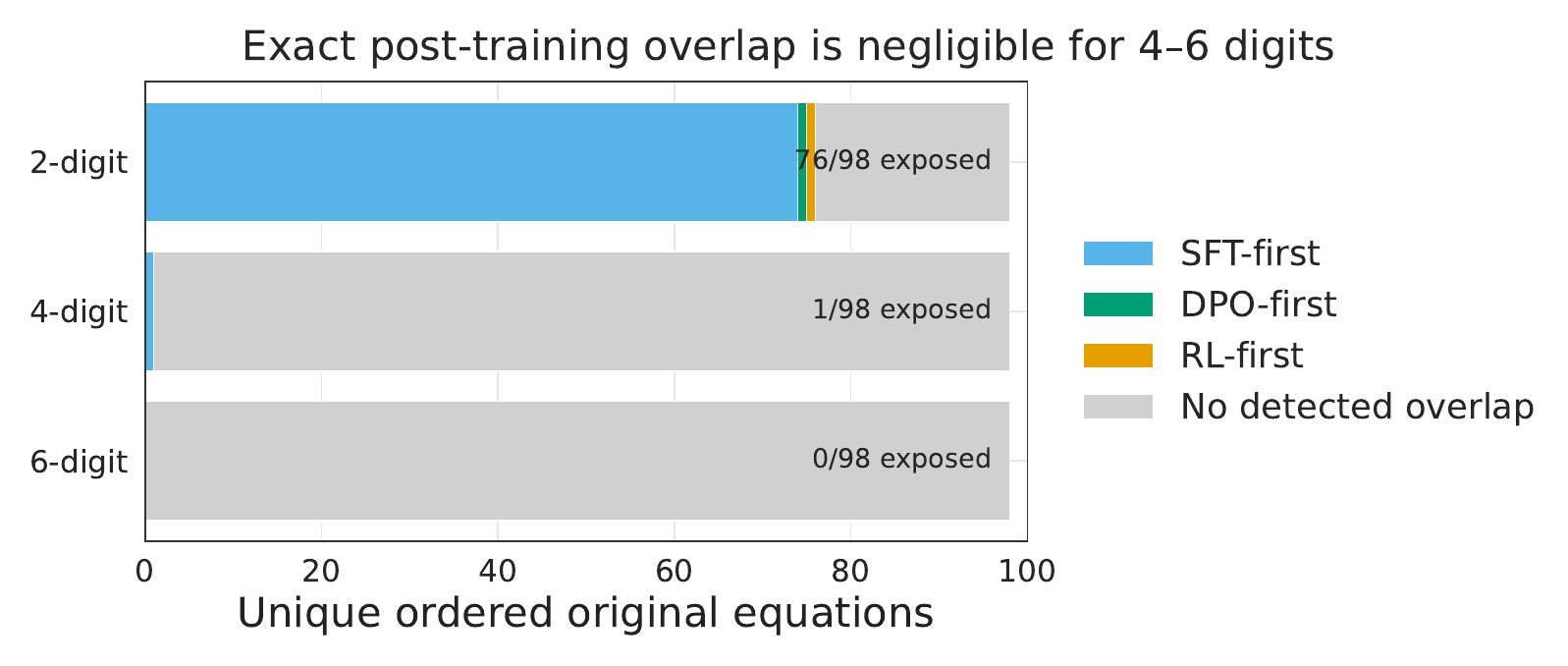}
    \caption{Exact symbolic overlap between the evaluation equations and the public post-training data. Each row contains 98 unique ordered equations after duplicate equations from different carry categories are collapsed. Coloured segments show the earliest post-training stage in which an exact relation occurs; grey denotes no detected occurrence. The search finds 76 two-digit equations, one four-digit equation, and no six-digit equations.}
    \label{fig:memorization-overlap-audit}
\end{figure}

We assess recall of individual evaluation equations with two complementary tests. First, we search the public post-training data released for the model
family for exact occurrences of the evaluation equations. Second, we compare each original equation with mathematically equivalent equations that have the same result and the same column-wise carry pattern.

\paragraph{Exact overlap with public post-training data.}
The overlap test uses the 384 standard-format, Answer Only equations from the behavioural evaluation: 128 each at 2, 4, and 6 digits. Because the same
equation can be sampled in more than one carry category, we collapse exact duplicates before searching. This leaves 98 unique ordered equations at each digit length. We search 2,581,998 records from the public supervised fine-tuning, preference-optimization, and reinforcement-learning collections.
An equation is counted as present when its exact symbolic addition relation appears in a model input or together with the expected answer. If it occurs in
more than one collection, it is assigned to the earliest post-training stage
in which it appears.

The search finds 76 of 98 two-digit equations, one of 98 four-digit equations, and no six-digit equations (\Cref{fig:memorization-overlap-audit}). The single four-digit match is \(9999 + 1 = 10000\). Exact overlap with these public post-training collections is therefore rare in the four- and six-digit evaluation sets. This test does not cover the model's pretraining data and does not test whether related examples teach reusable arithmetic patterns.

\section{Paired Comparison of Prompting Conditions}
\label{app:paired-prompt-modes}

\begin{table*}[t]
    \centering
    \small
    \setlength{\tabcolsep}{4pt}
    \begin{tabular}{lrrrrrrrr}
        \toprule
        Checkpoint & Matched \(n\) & Answer Only & Free CoT
        & \(\Delta\) [95\% CI] & Gains & Losses
        & Exact \(p\) & Holm \(p\) \\
        \midrule
        SFT  & 383 & 79.1 & 84.1 & $+5.0$ [0.0, 9.9]
             & 58 & 39 & .0671 & .1341 \\
        DPO  & 312 & 83.3 & 80.8 & $-$2.6 [$-$8.0, 2.6]
             & 32 & 40 & .4096 & .4096 \\
        Full & 291 & 86.9 & 93.1 & $+6.2$ [1.7, 10.7]
             & 32 & 14 & .0114 & .0341 \\
        \bottomrule
    \end{tabular}
    \caption{Paired comparison of Answer Only and Free CoT prompting among
    completed responses. Accuracies are percentages; \(\Delta\) is Free CoT
    minus Answer Only in percentage points, with an unadjusted 95\%
    problem-paired bootstrap confidence interval. Gains and losses count
    problems that Free CoT changed from incorrect to correct and from correct
    to incorrect, respectively. Holm \(p\) gives the exact McNemar
    \(p\)-value after correction across checkpoints.}
    \label{tab:paired-prompt-mode-results}
\end{table*}

We compared Answer Only and Free CoT prompting on matched arithmetic
problems. For each checkpoint, we retained only problems for which both
prompting conditions produced a completed response. This restriction measures
final-answer accuracy among completed responses, rather than treating a
failure to terminate as an incorrect answer. The analysis pooled 2-, 4-, and
6-digit addition, with each retained problem receiving equal weight. The
matched samples contained 128, 128, and 127 problems of each respective length for SFT; 127, 106, and 79 for DPO; and 128, 100, and 63 for Full.

For retained problem \(i\), let \(Y_{i,F}\) and \(Y_{i,A}\) indicate whether
the final answer was correct under Free CoT and Answer Only, respectively. We
estimated the prompt-mode effect as

\begin{equation}
    \widehat{\Delta}
    = \frac{1}{n}\sum_{i=1}^{n}\left(Y_{i,F}-Y_{i,A}\right).
    \label{eq:paired-prompt-effect}
\end{equation}

We obtained 95\% confidence intervals by resampling whole problem pairs with
replacement 10,000 times and taking the 2.5th and 97.5th percentiles of the
resulting mean differences. We also applied a two-sided exact McNemar test to
the paired binary outcomes for each checkpoint. The three resulting
\(p\)-values were adjusted using Holm's method, with adjusted \(p<.05\) as the significance threshold.

Free CoT changed final-answer accuracy by \(+5.0\) percentage points for SFT
(95\% CI \([0.0,9.9]\), adjusted \(p=.134\)), \(-2.6\) points for DPO
(95\% CI \([-8.0,2.6]\), adjusted \(p=.410\)), and \(+6.2\) points for Full
(95\% CI \([1.7,10.7]\), adjusted \(p=.034\)). Thus, only Full showed a
significant within-checkpoint improvement. Because the matched problem sets
differ across checkpoints, these tests do not establish that the change at
Full is significantly larger than the changes at SFT or DPO. We report
completion behavior separately because this analysis is conditional on both
responses completing. See \Cref{tab:paired-prompt-mode-results} for a summary of these results.

\subsection{Direct Comparison of the SFT and Full Prompt Effects}
\label{app:sft-full-prompt-interaction}

The preceding tests ask whether Free CoT changes accuracy within each
checkpoint. A significant result for Full and a non-significant result for SFT
do not, by themselves, show that the two prompt-mode effects differ. We
therefore tested this difference directly. To ensure that both checkpoint
effects were estimated from the same problems, we retained only problems for
which SFT and Full each produced completed responses under both Answer Only and
Free CoT prompting. This common sample contained 290 problems: 128 two-digit,
100 four-digit, and 62 six-digit problems.

Let \(Y_{i,m,p}\) indicate correctness for problem \(i\), checkpoint \(m\),
and prompt mode \(p\). We estimated the SFT--Full interaction as the difference
between their paired prompt-mode effects,

\begin{equation}
    \widehat{\Gamma}
    = \frac{1}{n}\sum_{i=1}^{n}
    \left[
        \left(Y_{i,\mathrm{Full},F}-Y_{i,\mathrm{Full},A}\right)
        - \left(Y_{i,\mathrm{SFT},F}-Y_{i,\mathrm{SFT},A}\right)
    \right].
    \label{eq:sft-full-prompt-interaction}
\end{equation}

We formed a 95\% confidence interval by resampling complete four-response
problem blocks with replacement 10,000 times and taking the 2.5th and 97.5th
percentiles of \(\widehat{\Gamma}\). For the hypothesis test, we used a
two-sided exact paired-permutation test. Under the null hypothesis that the
prompt-mode effect is exchangeable between SFT and Full, the SFT and Full
response pairs were swapped as a block within each problem. Because this was
one planned comparison, we used \(p<.05\) without a multiple-comparison
correction.

\begin{table}[t]
    \centering
    \small
    \setlength{\tabcolsep}{4pt}
    \begin{tabular}{lr}
        \toprule
        Quantity & Estimate \\
        \midrule
        Common problems & 290 \\
        SFT: Answer Only & 86.6\% \\
        SFT: Free CoT & 89.0\% \\
        SFT prompt effect & $+2.4$ pp \\
        Full: Answer Only & 86.9\% \\
        Full: Free CoT & 93.1\% \\
        Full prompt effect & $+6.2$ pp \\
        Full minus SFT interaction & $+3.8$ pp \\
        95\% paired-bootstrap CI & [$-$2.1, 9.7] pp \\
        Exact paired-permutation \(p\) & .255 \\
        \bottomrule
    \end{tabular}
    \caption{Direct comparison of the Free CoT-minus-Answer Only effect between Full and SFT. All quantities use the same problems, each with a completed response in all four checkpoint-by prompt conditions.}
    \label{tab:sft-full-prompt-interaction}
\end{table}

On this common sample, Free CoT increased accuracy by 2.4 percentage points
for SFT and 6.2 points for Full. The resulting Full-minus-SFT interaction was
\(+3.8\) points (95\% CI \([-2.1,9.7]\); exact paired-permutation \(p=.255\)).
The estimated gain was therefore larger for Full, but this analysis does not
establish that its prompt-mode effect differs significantly from SFT's. See \Cref{tab:sft-full-prompt-interaction} for a summary of these results.

\section{Additional Behavioural Results and Ablations}
\label{app:additional-results}

\begin{figure}[h]
    \centering
    \includegraphics[width=\linewidth]{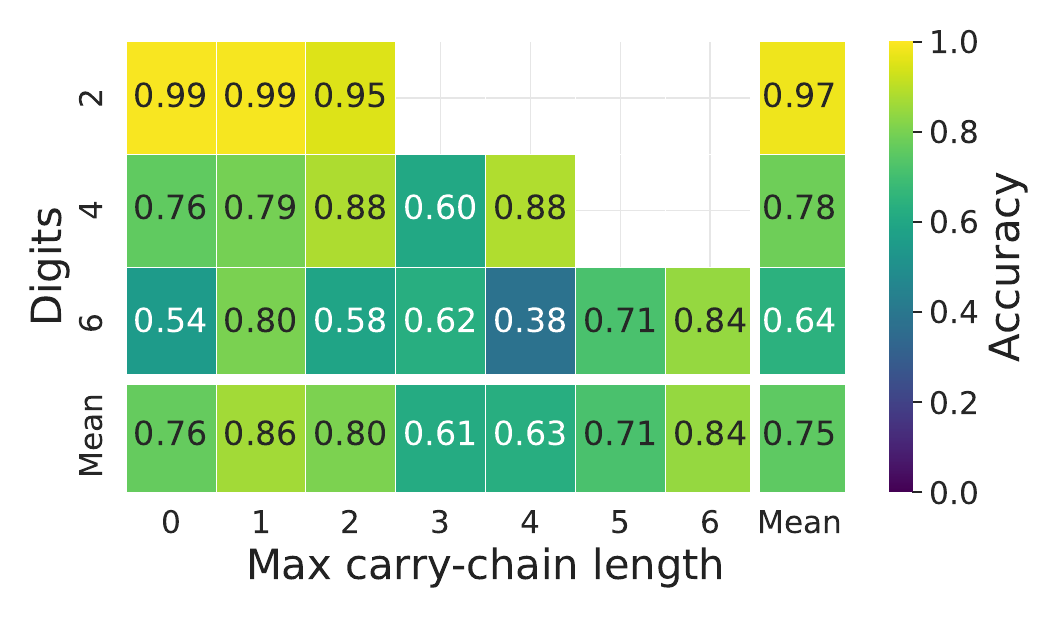}
    \caption{SFT final-answer accuracy by operand length and maximum
    carry-chain length. Each interior cell pools standard-format examples with
    the indicated properties across Answer Only and Free CoT. Blank cells are
    combinations absent from the evaluation set. The marginal row and column
    are unweighted means of the displayed interior cells.}
    \label{fig:carry-chain-accuracy-sft}
\end{figure}

\begin{figure}[h]
    \centering
    \includegraphics[width=\linewidth]{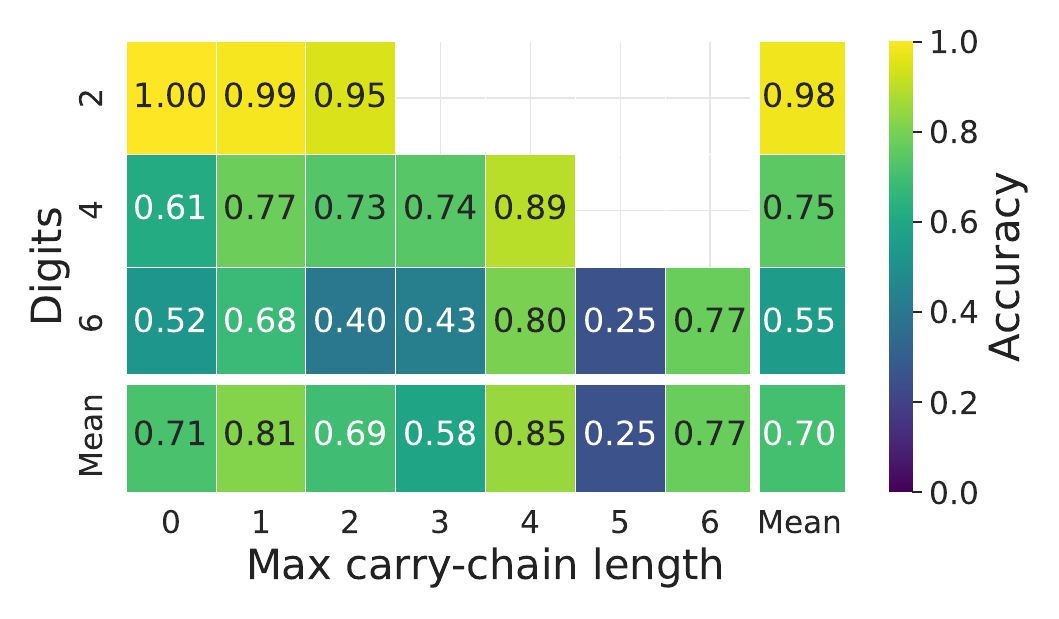}
    \caption{DPO final-answer accuracy by operand length and maximum
    carry-chain length. Each interior cell pools standard-format examples with
    the indicated properties across Answer Only and Free CoT. Blank cells are
    combinations absent from the evaluation set. The marginal row and column
    are unweighted means of the displayed interior cells.}
    \label{fig:carry-chain-accuracy-dpo}
\end{figure}

This section reports final-answer accuracy by carry structure, operand length, input format, and output format. Unless noted otherwise, results pool the Answer Only and Free CoT prompt instances and exclude responses that reach the 2,048-token limit without terminating.

The input- and output-format comparisons use separate arithmetic samples. The standard condition contains 384 problems, while each nonstandard condition contains 192 problems balanced across operand lengths. Every problem is evaluated under both prompt conditions, and no exact equation is shared across format samples. These comparisons therefore measure robustness to the full formatting condition; they are not paired estimates with arithmetic content held fixed.

\begin{figure}[h]
    \centering
    \includegraphics[width=\linewidth]{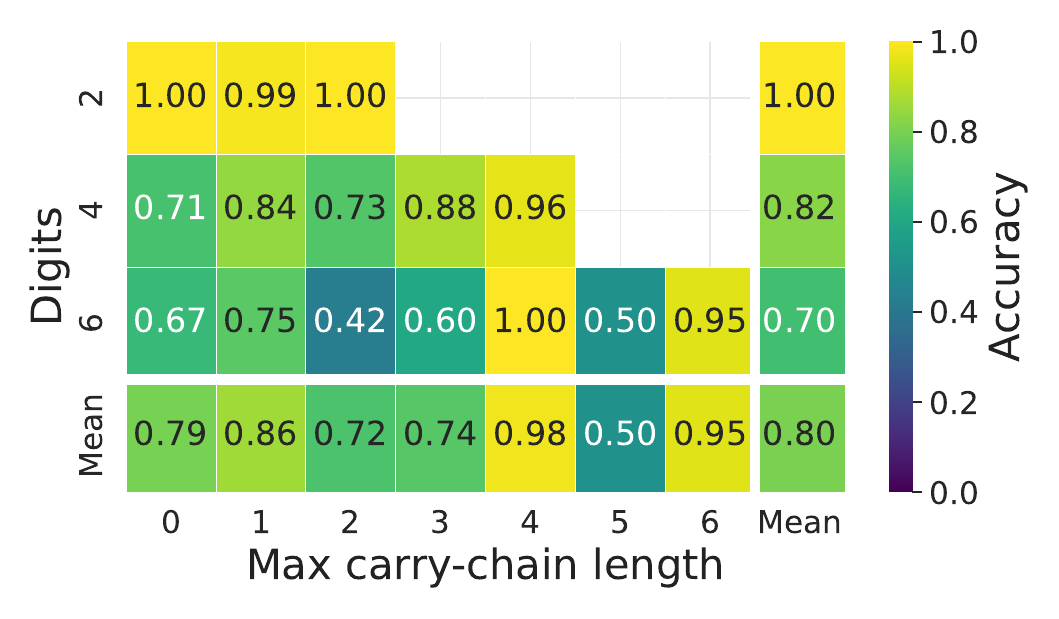}
    \caption{Full final-answer accuracy by operand length and maximum
    carry-chain length. Each interior cell pools standard-format examples with
    the indicated properties across Answer Only and Free CoT. Blank cells are
    combinations absent from the evaluation set. The marginal row and column
    are unweighted means of the displayed interior cells.}
    \label{fig:carry-chain-accuracy-full}
\end{figure}

\paragraph{Carry-chain length.}
The maximum carry-chain length is the longest sequence of adjacent columns in
which a carry propagates. \Cref{fig:carry-chain-accuracy-sft,fig:carry-chain-accuracy-dpo,fig:carry-chain-accuracy-full}
report accuracy jointly by this quantity and operand length. When the
displayed carry-chain cells are weighted equally, SFT obtains accuracies of
0.97, 0.78, and 0.64 at 2, 4, and 6 digits; DPO obtains 0.98, 0.75, and 0.55;
and Full obtains 1.00, 0.82, and 0.70. Accuracy generally decreases with
operand length, but it is not monotonic in carry-chain length within each
digit stratum. Some constructed long-chain problems are easier than the more
varied problems with shorter chains.

\begin{figure}[h]
    \centering
    \includegraphics[width=\linewidth]{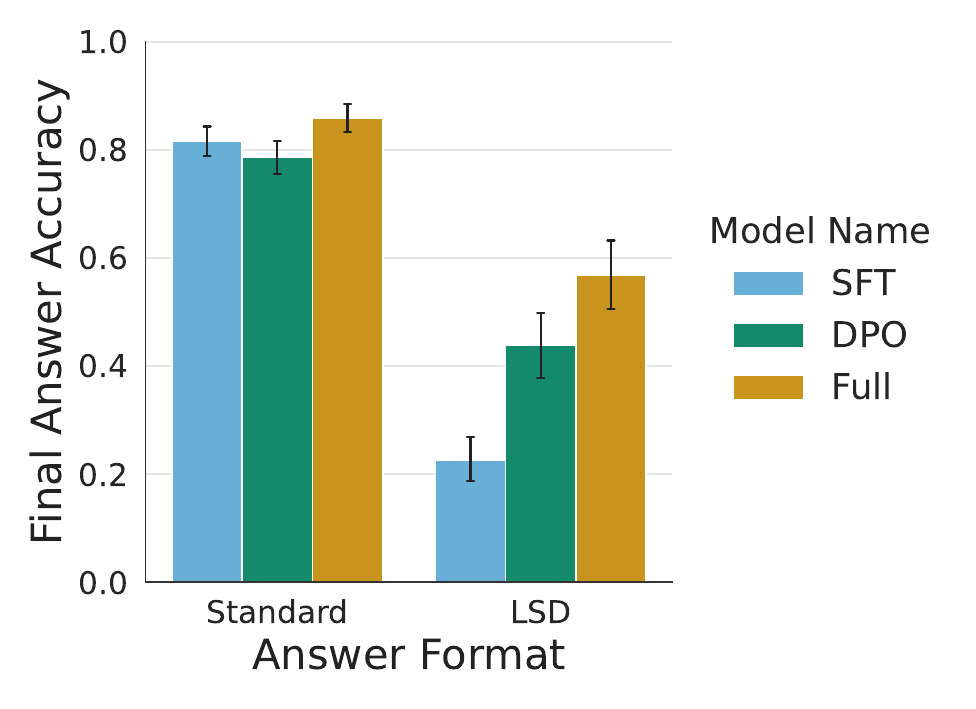}
    \caption{Final-answer accuracy with the standard output convention and
    with the result digits written from least to most significant (LSD).
    Operands are conventionally formatted in both conditions. The standard
    and LSD conditions use separate samples of 384 and 192 problems,
    respectively, with both prompt conditions for every problem. Error bars
    are 95\% bootstrap confidence intervals.}
    \label{fig:answer-format-ablation}
\end{figure}

\paragraph{Answer format.}
The standard condition requires the usual left-to-right result. In the
least-significant-digit-first (LSD) condition, the operands remain
conventional but the result must be written from right to left. Accuracy is
lower on the LSD sample for all three checkpoints
(\Cref{fig:answer-format-ablation}): SFT decreases from 81.6\% to 22.7\%, DPO
from 78.7\% to 43.8\%, and Full from 85.9\% to 56.9\%. Because the two samples
contain different equations, these differences include both the change in
answer order and sampling variation.

\begin{figure}[h]
    \centering
    \includegraphics[width=\linewidth]{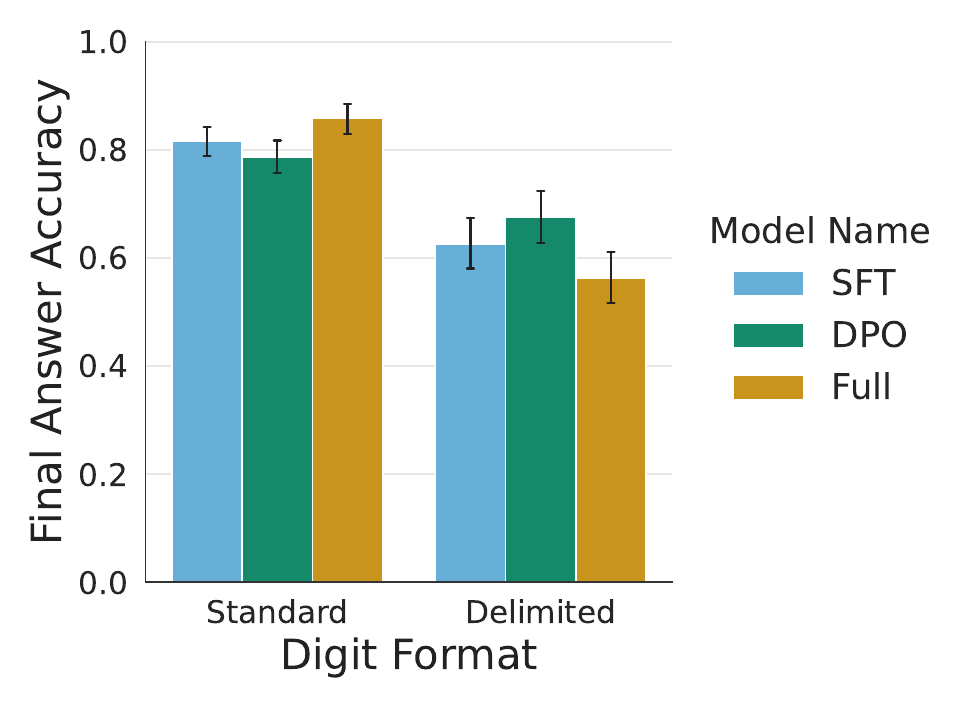}
    \caption{Final-answer accuracy with conventionally written operands and
    with \texttt{|} between adjacent operand digits. The answer remains
    conventionally formatted, and the delimited prompt explains the notation.
    The standard and delimited conditions use separate samples of 384 and 192
    problems, respectively, with both prompt conditions for every problem.
    Error bars are 95\% bootstrap confidence intervals.}
    \label{fig:digit-format-ablation}
\end{figure}

\paragraph{Operand format.}
The standard condition writes each operand conventionally. The delimited
condition places \texttt{|} between adjacent digits and explains the notation
in the prompt; the answer remains conventionally formatted. Accuracy decreases
from 81.6\% to 62.5\% for SFT, from 78.7\% to 67.5\% for DPO, and from 85.9\%
to 56.3\% for Full (\Cref{fig:digit-format-ablation}). As in the answer-format
comparison, the standard and delimited samples contain different equations.

\begin{figure}[h]
    \centering
    \includegraphics[width=\linewidth]{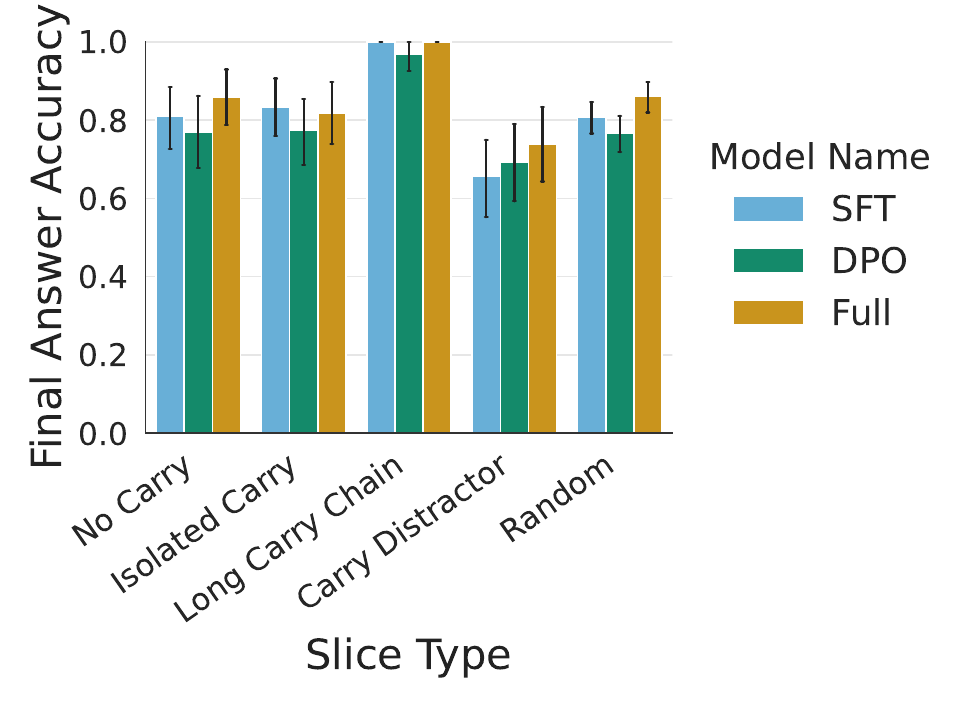}
    \caption{Final-answer accuracy across controlled carry structures. No
    Carry has no carry-producing column; Isolated Carry has one
    non-propagating carry; Long Carry Chain propagates a carry across every
    operand digit; Carry Distractor has locally carry-suggestive patterns but
    no long chain; and Random provides carry-balanced coverage. All examples
    use standard operand and answer formats. Bars pool both prompt conditions;
    error bars are 95\% bootstrap confidence intervals.}
    \label{fig:carry-slice-ablation}
\end{figure}

\paragraph{Carry structure.}
The controlled slices contain problems with no carry, one isolated carry, a
carry propagated across every operand digit, or carry-suggestive digit patterns
that do not form a long chain. A random slice provides broader coverage. The
long-chain slice has the highest accuracy for all three checkpoints
(96.8--100\%), whereas the carry-distractor slice has the lowest
(65.6--73.8\%; \Cref{fig:carry-slice-ablation}). Carry count and chain length
therefore do not provide a complete ordering of problem difficulty.

\begin{figure}[h]
    \centering
    \includegraphics[width=\linewidth]{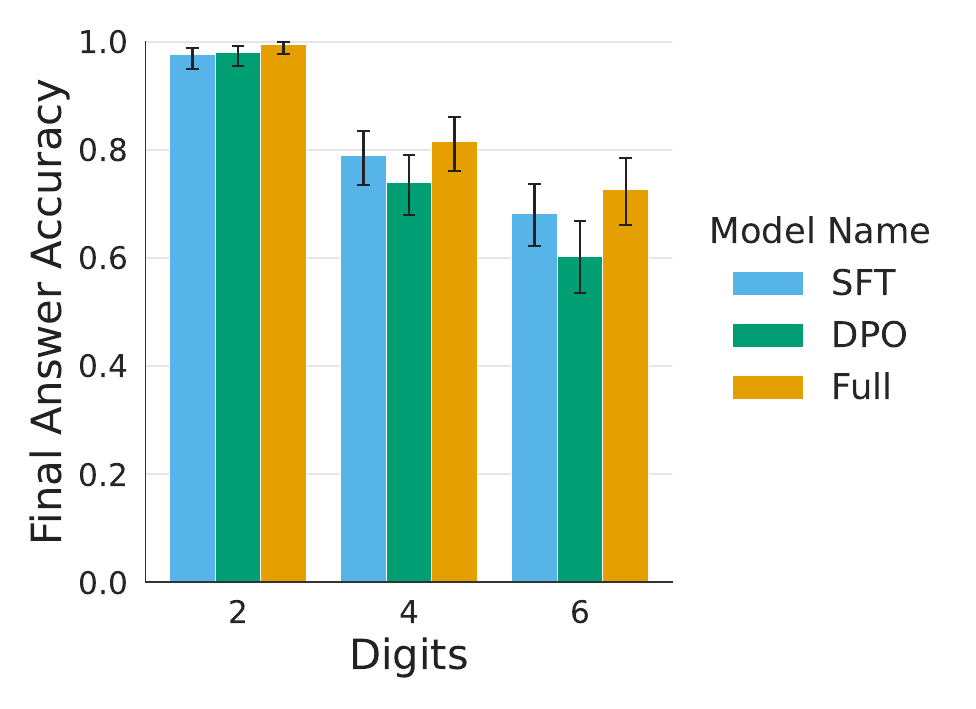}
    \caption{Final-answer accuracy by operand length for standard-format inputs and answers. Bars pool Answer Only and Free CoT prompt instances; error bars are 95\% Wilson confidence intervals for the pooled binomial proportions.}
    \label{fig:standard-accuracy-by-digits}
\end{figure}

\paragraph{Operand length.}
\Cref{fig:standard-accuracy-by-digits} pools all standard-format examples at
each operand length. Accuracy is 97.7--99.6\% at two digits and decreases at
four and six digits. At six digits, SFT, DPO, and Full obtain 68.2\%, 60.4\%,
and 72.8\%, respectively. These estimates weight every eligible prompt
instance equally, unlike the heatmap margins, which weight each displayed
carry-chain cell equally.

\section{Probe Training Procedure and Results}
\label{app:probes}

We use linear probes to test whether the residual stream contains information
about the intermediate variables of column-wise addition. The probes are
diagnostic: successful classification establishes that a variable can be
recovered by a linear decision boundary, but not that the model uses that
boundary when producing its answer. We train probes for SFT and Full, which
represent the endpoints of the post-training sequence considered in the main
text.

\subsection{Targets and Activation Collection}

We index columns from right to left, so column $i=0$ is the units column.
For operand digits $a_i$ and $b_i$, let $c_i$ be the carry entering
column $i$, with $c_0=0$. The column computation is

\begin{align}
    r_i &= a_i + b_i + c_i, \\
    y_i &= r_i \bmod 10, \\
    c_{i+1} &= \left\lfloor r_i / 10 \right\rfloor .
\end{align}

Separate probes predict the incoming carry $c_i$, the outgoing carry
$c_{i+1}$, and the result digit $y_i$. Carry prediction is binary and
result-digit prediction is ten-way classification. A distinct probe is fitted
for every checkpoint, target, arithmetic column, token-location type, and
transformer layer. Reported profiles aggregate the resulting test accuracies
over arithmetic columns.

The dataset contains disjoint training and test sets with 64 problems at each
of 2, 4, and 6 digits. Carry counts are balanced within each split. Every
problem is presented under both Answer Only and Free CoT, but the results in
the main text use Free CoT. Reasoning traces are sampled with temperature
0.6, nucleus probability 0.95, and a maximum of 2,048 generated tokens.
Responses that reach this limit without terminating are excluded before probe
training. All SFT responses terminate within the limit. Full retains every
two-digit response but fewer of the longer responses, as shown in
\Cref{tab:probe-sample-counts}.

\begin{table}[t]
    \centering
    \small
    \setlength{\tabcolsep}{3.5pt}
    \begin{tabular}{@{}llrrrr@{}}
        \toprule
        Checkpoint & Split & 2-digit & 4-digit & 6-digit & Total \\
        \midrule
        SFT  & Train & 64 & 64 & 64 & 192 \\
        SFT  & Test  & 64 & 64 & 64 & 192 \\
        Full & Train & 64 & 44 & 17 & 125 \\
        Full & Test  & 64 & 36 & 16 & 116 \\
        \bottomrule
    \end{tabular}
    \caption{Free CoT problems retained for probe training and evaluation
    after excluding responses that do not terminate within 2,048 tokens. The
    original training and test sets each contain 64 problems at every digit
    length for each checkpoint.}
    \label{tab:probe-sample-counts}
\end{table}

We collect the residual stream after each of the model's 32 transformer
layers. The full set of token locations comprises the operand digits and
question mark in the prompt; the final prompt token; the first, one-third,
two-thirds, and final tokens of the generated reasoning; and the tokens
containing the final-answer digits. Fractions of the reasoning trace are
defined by generated-token count before the final answer begins. The main
profile uses three representative locations: the final prompt token, the
two-thirds reasoning token, and the answer-digit tokens.

\begin{table*}[t]
    \centering
    \small
    \setlength{\tabcolsep}{5pt}
    \begin{tabular}{@{}llccc@{}}
        \toprule
        Target & Token location & Majority baseline & SFT & Full \\
        \midrule
        Incoming carry & Prompt final  & 57.3 & 71.1 [65.5, 77.4] & 73.8 [69.1, 76.8] \\
        Incoming carry & CoT two-thirds & 57.3 & 57.2 [54.0, 61.5] & 60.8 [52.7, 66.8] \\
        Incoming carry & Answer digits & 57.0 & 66.9 [61.2, 70.2] & 63.0 [56.8, 66.3] \\
        \addlinespace
        Outgoing carry & Prompt final  & 54.8 & 72.2 [69.0, 76.6] & 72.7 [69.6, 75.3] \\
        Outgoing carry & CoT two-thirds & 54.8 & 58.0 [53.0, 62.5] & 57.1 [52.4, 61.9] \\
        Outgoing carry & Answer digits & 54.8 & 74.9 [65.5, 78.1] & 72.2 [66.4, 75.3] \\
        \addlinespace
        Result digit & Prompt final  & 25.2 & 22.8 [20.1, 25.9] & 22.7 [19.2, 24.2] \\
        Result digit & CoT two-thirds & 25.2 & 19.4 [15.1, 25.2] & 20.6 [17.0, 24.7] \\
        \bottomrule
    \end{tabular}
    \caption{Probe test accuracy on the test problems available for both SFT
    and Full, in percent. Each checkpoint entry first pools target columns in
    proportion to their number of test predictions and then averages equally
    over the 32 layers; brackets give the minimum and maximum layer accuracy.
    The majority baseline is computed under the same target-column weighting.
    Result-digit prediction at answer-digit locations is omitted because the
    target digit has already been emitted.}
    \label{tab:probe-summary}
\end{table*}

At prompt and reasoning locations, each problem contributes one activation to
each probe for a given target column. At the answer, a problem can contribute
several activations. For each target column, one probe is trained jointly on
all answer-digit locations; we do not fit a separate probe for each emitted
position. A tokenizer token can contain several decimal characters, in which
case those character positions share the same residual state. Predicting a
result digit from a token that already contains that digit would expose the
label directly, while merged tokens can associate the same state with several
digit labels. We therefore omit result-digit probes at answer-digit locations
from the main analysis. Answer-digit locations are retained for the two carry
targets, for which the emitted digit is not itself the label.

\subsection{Probe Fitting and Evaluation}

For every probe, each residual-stream feature is standardized using the
training split. We then fit an $L_2$-regularized logistic regression model
with inverse regularization strength $C=1$ and class-balanced training
weights. Binary logistic regression is used for the carry targets and
multinomial logistic regression for the result digit. We train a total of 7,744 fitted classifiers across the two checkpoints. The reported metric is accuracy on the held-out problem split. Hyperparameters are fixed before evaluation, and the main profile reports every layer rather than selecting a best layer on the test set.

\subsection{Probe Results}

The profiles in \Cref{fig:probe-profile} are similar across layers and between
SFT and Full. \Cref{tab:probe-summary} provides a numerical summary restricted
to the 116 test problems with a terminating Free CoT response from both checkpoints. Incoming- and outgoing-carry accuracies are highest at the final prompt token and at answer-digit locations. Accuracy at two-thirds of the reasoning trace is lower, but remains close across checkpoints. Averaged over layers, the largest SFT--Full difference among the carry entries in the table is 3.9 percentage points.

The result-digit probes behave differently. Their mean accuracies are
19.4--22.8\%, compared with a 25.2\% majority-class baseline. We therefore do not find evidence that a single linear probe
reliably recovers the eventual result digit at the final prompt token or at
two-thirds of the reasoning trace. The stronger carry results show linear
decodability, but they do not establish that the decoded direction is a causal
control variable.

\begin{table}[t]
    \centering
    \small
    \setlength{\tabcolsep}{2.5pt}
    \begin{tabular}{@{}lrrc@{}}
        \toprule
        Checkpoint & Problems & Contrasts & Mean [95\% CI] \\
        \midrule
        SFT  & 73 & 518 & $0.004\;[-0.003,\,0.012]$ \\
        Full & 73 & 522 & $-0.002\;[-0.009,\,0.005]$ \\
        \bottomrule
    \end{tabular}
    \caption{Problem-weighted effect of steering along the incoming-carry
    probe direction relative to three norm-matched orthogonal directions. The
    outcome is the change in counterfactual-versus-factual complete-answer log
    probability, measured in nats. Confidence intervals are 95\% percentile
    intervals from resampling whole arithmetic problems.}
    \label{tab:probe-steering-results}
\end{table}

\subsection{Steering with Probe Directions}
\label{app:probe-steering}

To test whether a decoded carry direction also controls the computation, we
intervene along the incoming-carry probe direction. Let $u$ be the normalized
probe direction in the original residual-stream coordinates, and let
$\mu_0$ and $\mu_1$ be the mean training-set projections onto $u$ for
examples without and with an incoming carry. For an activation $h$ whose
factual carry is $c\in\{0,1\}$, we clamp its projection to the opposite-class
mean:

\begin{equation}
    h' = h + \left(\mu_{1-c} - h^\top u\right)u.
    \label{eq:probe-direction-clamp}
\end{equation}

The intervention is applied at one token and one layer. We test layers 16 and
24 at the final prompt token, the two-thirds reasoning token, and the final
reasoning token. Evaluation uses 73 two- and four-digit problems that both
checkpoints answer correctly and for which toggling the selected incoming
carry changes one result digit without changing the outgoing carry. This
defines a counterfactual answer while leaving every other result digit fixed.

For each probe direction, we construct three random unit directions orthogonal
to $u$. The controls receive the same displacement magnitude as the carry
intervention. Let $y$ and $y^{\mathrm{cf}}$ denote the factual and
counterfactual complete answers, respectively, and define the model's
counterfactual preference at the intervention site as

\begin{equation}
    q(h) = \log p(y^{\mathrm{cf}} \mid h) - \log p(y \mid h),
\end{equation}

where each term is the summed teacher-forced log probability of the answer
tokens. The effect for a direction is $q(h')-q(h)$. Our primary contrast
subtracts the mean effect of the three orthogonal controls from the effect of
the carry direction, and then averages first within each arithmetic problem
and equally across problems.

We find that the
carry direction does not increase preference for the counterfactual answer
relative to the matched controls. The mean contrast is 0.004 for SFT and
$-0.002$ for Full, and both problem-bootstrap intervals include zero
(\Cref{tab:probe-steering-results}).

This is a negative result for a single-token, single-layer linear
intervention. It does not show that carry information is absent or causally
irrelevant; rather, the linearly decodable direction is not by itself a
sufficient control axis for the model's answer. A carry mechanism distributed
across tokens, layers, or nonlinear features would not be isolated by this test.

\section{Negative Results on Causal Interventions}
\label{app:causal-interventions}

\begin{table*}[t]
    \centering
    \small
    \setlength{\tabcolsep}{4.5pt}
    \begin{tabular}{@{}llrrrcc@{}}
        \toprule
        Receiver & Reasoning source
        & \multicolumn{3}{c}{Correct-answer log probability}
        & \multicolumn{2}{c}{Final-answer accuracy (\%)} \\
        \cmidrule(lr){3-5}\cmidrule(l){6-7}
        & & One-third & Two-thirds & Complete & Two-thirds & Complete \\
        \midrule
        SFT  & SFT  & $-3.672$ & $-6.833$ & $-1.177$ & 19 & 89 \\
        SFT  & Full & $-4.678$ & $-6.229$ & $-0.915$ & 22 & 96 \\
        Full & SFT  & $-6.355$ & $-9.882$ & $-1.199$ & 20 & 91 \\
        Full & Full & $-8.001$ & $-10.162$ & $-1.003$ & 17 & 95 \\
        \bottomrule
    \end{tabular}
    \caption{Cross-checkpoint replay results on 100 arithmetic problems.
    Log probabilities are sums over the correct answer tokens, averaged over
    problems. Accuracies are from deterministic continuations. Without
    supplied reasoning, SFT and Full obtain 99\% and 98\% accuracy,
    respectively; their mean correct-answer log probabilities are $-0.154$
    and $-0.392$.}
    \label{tab:natural-replay-results}
\end{table*}

This section reports three interventions that did not support a simple,
checkpoint-specific causal account of carry computation. First, we replay
natural reasoning traces across checkpoints to test whether later
post-training changes how the model uses a supplied chain of thought. We then
test two activation-level interventions: a learned multidimensional carry
subspace and direct transfer of short residual-stream spans. These experiments
are informative about the particular intervention tested, but a null result
does not show that carry information is absent from the network.

\subsection{Cross-Checkpoint Replay of Natural Reasoning}

We begin with 100 addition problems: 64 with two-digit operands and 36 with four-digit operands. SFT and Full independently generate a Free CoT response for every problem. The SFT traces end in the correct answer on 90 problems, and the Full traces do so on 97. We retain all traces rather than conditioning the crossover analysis on source correctness.

For each problem, a source trace is supplied by teacher forcing to either the SFT or Full receiver. We evaluate prefixes ending at one-third and two thirds of the reasoning and immediately before the source's final answer. At the two partial locations, an answer cue is appended before scoring or generation; at the final location, the receiver continues directly from the completed reasoning. The no-reasoning condition scores the answer directly after the prompt. At every location we measure the summed log probability of the correct answer tokens. We additionally decode greedily from the no-reasoning, two-thirds, and final locations.

\Cref{tab:natural-replay-results} shows that the identity of the supplied reasoning matters, particularly once the reasoning is complete, but the two receivers respond similarly to it. At the final location, replacing an SFT trace with a Full trace improves exact accuracy by 7 percentage points for the SFT receiver and by 4 points for the Full receiver. The resulting receiver-by-source interaction is $-3$ points (95\% problem-bootstrap CI $[-8,1]$). For the teacher-forced answer score, define

\begin{equation}
    L_{r,s} = \log p_r(y \mid x,z_s),
\end{equation}

where $r$ denotes the receiver, $s$ the source checkpoint, and $z_s$ the
complete source reasoning. The corresponding interaction,

\begin{equation}
    (L_{\mathrm{Full},\mathrm{Full}}-L_{\mathrm{Full},\mathrm{SFT}})
    - (L_{\mathrm{SFT},\mathrm{Full}}-L_{\mathrm{SFT},\mathrm{SFT}}),
\end{equation}

is $-0.066$ nats (95\% CI $[-0.278,0.167]$).  Thus, Full-sourced reasoning
outperforms SFT-sourced reasoning in this comparison, but we do not detect a
receiver-by-source interaction.

The incorrect source traces provide a stronger manipulation check. There are
10 problems for which the SFT trace has a clean but incorrect final answer
while both receivers answer correctly without supplied reasoning. After the
complete incorrect trace is replayed, the SFT receiver reproduces the source's wrong answer on all 10 problems. The Full receiver reproduces it on seven and recovers the correct answer on three. Natural reasoning therefore has a clear causal effect on the output. The negative result is narrower: we find no reliable evidence that later post-training changes the receiver's sensitivity to the source checkpoint.

\subsection{Distributed Alignment Search}
\label{app:das}

A linear probe may miss a variable encoded in a rotated, multidimensional
subspace. We therefore train a Boundless Distributed Alignment Search
intervention \citep{wu2024interpretabilityscaleidentifyingcausal} separately
for SFT and Full. The model weights are frozen while
an orthogonal rotation and a continuous subspace boundary are learned so that
swapping the selected component between two activations should exchange the
incoming carry.

Training uses four-digit problem pairs in which the carry entering the tens or
hundreds column differs between the pair. The preceding output digit is
matched, the carry leaving the target column is unchanged, and the
counterfactual answer differs from the factual answer at only the target
digit. The data contain 1,024 training pairs, 256 tuning pairs, and 64 held-out
evaluation pairs. Target column and swap direction are balanced. Candidate
interventions are screened at layers 7, 15, 23, and 31, both after the
preceding column and before the target column. For the selected layer and
location, three optimization seeds are compared on the tuning set before the
intervention is frozen.

For both checkpoints, the selected intervention is at layer 7 immediately
before the target-column computation. The learned regions occupy about half
of the 4,096-dimensional residual stream. Despite their size, neither
intervention transfers reliably to the held-out controlled problems
(\Cref{tab:boundless-das-results}). Interchange accuracy is 1.6\% for SFT and
12.5\% for Full. Off-target digit preservation is also not satisfactory.

\begin{table}[t]
    \centering
    \small
    \setlength{\tabcolsep}{5pt}
    \begin{tabular}{@{}lrr@{}}
        \toprule
        Measure & SFT & Full \\
        \midrule
        Subspace dimensions & 2,137 (52.2\%) & 2,102 (51.3\%) \\
        Tuning interchange & 3.1\% & 9.4\% \\
        Held-out interchange & 1.6\% & 12.5\% \\
        Target digit changed & 1.6\% & 14.1\% \\
        Other digits preserved & 62.5\% & 90.3\% \\
        \bottomrule
    \end{tabular}
    \caption{Performance of the selected distributed carry interventions. Interchange is the rate at which the generated answer changes to the exact counterfactual answer. Other digits preserved is the rate at which all non-target answer digits remain unchanged. The held-out evaluation contains 64 pairs per checkpoint.}
    \label{tab:boundless-das-results}
\end{table}

\subsection{Cross-Checkpoint Activation Patching}
\label{app:activation-patching}

We next test a more direct activation-patching intervention
\citep{heimersheim2024useinterpretactivationpatching} that does not require
learning a subspace. At a decoder-layer boundary, the earlier computation acts as a
\textit{writer} and the remaining layers as a \textit{reader}. For a matched
pair with opposite incoming carries, we replace a short span at the input to
the reader with the corresponding residual-stream states produced by the
writer. Crossing SFT and Full yields four writer--reader combinations. A
same-input transfer controls for disruption caused merely by crossing
checkpoints.

Let $m(h)$ be the receiver's log-probability margin between the locally
counterfactual and factual target digit after receiving states $h$. We define
the carry effect as

\begin{equation}
    \Delta_{\mathrm{carry}}
    = m(h_{\mathrm{opposite\ carry}})
    - m(h_{\mathrm{same\ input}}).
    \label{eq:cross-patch-carry-effect}
\end{equation}

A positive value means that states from the opposite-carry problem move the
receiver toward the digit predicted by the carry swap. We compute the same
contrast using the summed log-probability margin of the complete answer.

We screen all 31 non-embedding layer boundaries and spans of one, two, and eight tokens on 16 controlled problems. Configurations are ranked by the
weaker of their SFT-to-SFT and Full-to-Full target effects, after which two distinct layer boundaries are retained. Both selected configurations replace
the final eight tokens before the target-column computation, at layers 27 and 28. On a separate set of 64 controlled problems, the primary layer-27
configuration has mean target effects of $-0.006$ nats for SFT and $-0.016$ nats for Full. The patch favors the counterfactual target digit in 42.2\% of
cases for both checkpoints and never produces the complete predicted answer. Although 92.2\% of cases preserve all other answer digits, the target effect,
preference rate, and exact-answer rate all fail their prespecified checks.

We nevertheless record local teacher-forced scores on a separately generated
natural-reasoning sample, while omitting autoregressive continuations. Of 128
fixed matched problem pairs, 15 contain complete traces with an unambiguous
boundary in both members: 10 pairs written by SFT and five by Full. Both swap
directions are evaluated, giving 30 directed cases per writer--reader cell.
The pooled results in \Cref{tab:natural-cross-patch-results} show positive
local target effects in every cell. However, the complete-answer effects are
smaller, and the writer--reader interaction is close to zero at both layer
boundaries. These estimates are exploratory: the controlled check failed and
the natural sample is small after boundary matching.

\begin{table}[t]
    \centering
    \scriptsize
    \setlength{\tabcolsep}{2.2pt}
    \begin{tabular}{@{}clcc@{}}
        \toprule
        Layer & Writer $\to$ reader & Target [95\% CI] & Answer [95\% CI] \\
        \midrule
        27 & SFT $\to$ SFT   & $0.29\;[0.09,0.50]$ & $0.02\;[-0.07,0.14]$ \\
        27 & SFT $\to$ Full  & $0.38\;[0.19,0.55]$ & $0.05\;[-0.03,0.14]$ \\
        27 & Full $\to$ SFT  & $0.21\;[0.00,0.42]$ & $0.06\;[-0.05,0.19]$ \\
        27 & Full $\to$ Full & $0.27\;[0.05,0.47]$ & $0.10\;[-0.05,0.30]$ \\
        \addlinespace
        28 & SFT $\to$ SFT   & $0.28\;[0.10,0.47]$ & $0.03\;[-0.03,0.11]$ \\
        28 & SFT $\to$ Full  & $0.33\;[0.17,0.49]$ & $0.08\;[0.02,0.15]$ \\
        28 & Full $\to$ SFT  & $0.24\;[0.07,0.42]$ & $0.08\;[-0.03,0.21]$ \\
        28 & Full $\to$ Full & $0.23\;[0.05,0.40]$ & $0.13\;[0.01,0.30]$ \\
        \bottomrule
    \end{tabular}
    \caption{Exploratory activation-patching effects on 15 eligible natural
    problem pairs. Effects are changes in counterfactual-versus-factual log
    probability, in nats, relative to the same-input cross-checkpoint control.
    Each cell contains 30 directed cases. Confidence intervals are 95\%
    percentile intervals from resampling whole matched problem pairs.}
    \label{tab:natural-cross-patch-results}
\end{table}

The checkpoint comparisons do not indicate a changed carry code. At layer 27,
the writer--reader interaction is $-0.025$ nats for the target digit (95\% CI
$[-0.084,0.030]$) and $-0.001$ for the complete answer
($[-0.074,0.077]$). At layer 28, the corresponding interactions are $-0.058$
($[-0.126,0.005]$) and $0.002$ ($[-0.059,0.063]$). Likewise, the Full-to-Full
minus SFT-to-SFT target effect is $-0.022$ nats at layer 27
($[-0.267,0.173]$) and $-0.054$ at layer 28 ($[-0.263,0.114]$).

Together, these results do not support a simple interpretation in which
post-SFT training introduces a carry variable that is easily isolated by the tested interventions and has checkpoint-specific writer--reader compatibility. They do not rule out a distributed mechanism that spans more tokens or layers, depends on non-linear features, or changes with the model's reasoning strategy.

\section{Answer Emergence and Commitment During the CoT}
\label{app:answer-emergence}

We distinguish two questions about an intermediate reasoning prefix. First, can the model produce the correct final answer when asked at that point? Second, has the answer become stable, in the sense that it remains correct at every later observed boundary? The first measures immediate answer availability while the second measures observed commitment.

\subsection{Evaluation at Natural Reasoning Boundaries}

The analysis uses the 45 four-digit and 32 six-digit problems for which SFT, DPO, and Full all produce a usable Free CoT trace. This gives 231 source traces. Each trace is divided into natural reasoning units, such as headings, sentences, and individual arithmetic steps. A boundary's normalized progress is the number of source-reasoning tokens observed through that boundary, divided by the total number of reasoning tokens in the trace. The final prompt token is assigned progress zero and the end of the reasoning is assigned progress one.

At every boundary, the source prefix is supplied by teacher forcing to each of the three receivers. We append the same final-answer cue and decode deterministically. Intermediate queries receive a 16-token answer budget; the endpoint query is allowed up to 64 tokens so that a complete response can terminate naturally. Essentially, this experiment supplies a portion of the CoT and then cuts it off by immediately asking for an answer at evenly spaced portions throughout the CoT. We parse the resulting final answer and compare it with the exact arithmetic result. To avoid treating a resumed calculation as a direct answer, continuations that don't immediately answer (e.g. by producing additional reasoning) are excluded.

For comparison across traces of different lengths, we evaluate the observed boundary nearest each tenth of normalized progress. Confidence bands are 95\% percentile intervals obtained by resampling arithmetic problems.

\begin{figure*}[t]
    \centering
    \includegraphics[width=\linewidth]{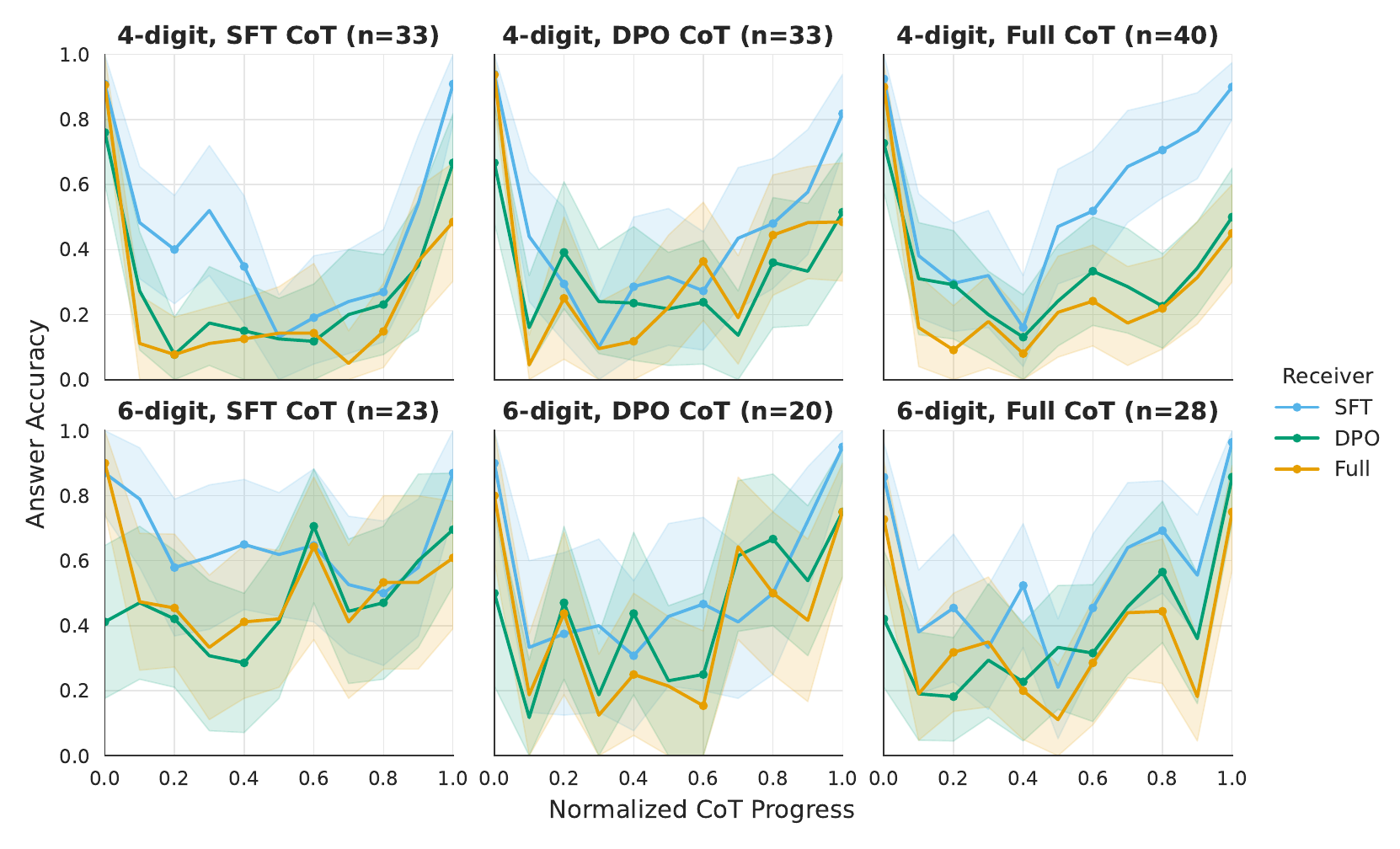}
    \caption{Immediate final-answer accuracy over normalized CoT progress.
    Rows show operand length, columns show the checkpoint that produced the
    reasoning, and line colour shows the receiver checkpoint. Each source
    cohort contains completed traces whose original final answer is correct;
    the panel title gives its size. At each tenth of progress, the nearest
    observed reasoning boundary is supplied to the receiver before requesting
    a direct answer. Non-direct intermediate continuations are excluded and
    completed endpoint decodes are retained. Shaded regions are 95\%
    problem-bootstrap confidence intervals.}
    \label{fig:answer-emergence-trajectory}
\end{figure*}

\subsection{A U-Shaped Answer Trajectory}

\Cref{fig:answer-emergence-trajectory} restricts attention to source traces
whose original final answer is correct. The six source cohorts contain 33,
33, and 40 four-digit traces from SFT, DPO, and Full, respectively, and 23,
20, and 28 six-digit traces. At the prompt boundary, many receivers can
already produce the answer. Accuracy then falls sharply after the initial
reasoning units, remains low or unstable through much of the trace, and rises
again near the endpoint. This qualitative U-shape appears across source
checkpoints, receiver checkpoints, and both operand lengths, although its
depth and recovery vary across panels. We note that the CoT progression does not include the final reasoning span which often includes the answer, which is why the answer accuracy does not perfectly equal 1 when the CoT progression is equal to 1.

The early decrease should not be interpreted as literal deletion of a known
answer. The measurement interrupts generation and asks for an answer at a
point where the visible reasoning may contain a partial decomposition, a
local column result, or an unfinished verification. Such prefixes can shift
the receiver's native output distribution toward intermediate quantities even
when information sufficient for the final answer remains elsewhere in the
residual stream. The late recovery is similarly compatible with several
processes: completing the arithmetic, integrating previously stated partial
results, recomputing them, or verifying an answer that was available earlier.
The experiment establishes that immediate answerability is non-monotonic; it
does not by itself identify which of these processes causes the recovery.

\subsection{Stable Commitment Occurs Late}

For a source trace and receiver, let $A_k$ indicate whether the direct answer
at observed boundary $k$ is correct. We define the observed commitment
boundary as

\begin{equation}
    k^{*} = \min\{k : A_j = 1 \text{ for every retained } j \geq k\}.
    \label{eq:observed-answer-commitment}
\end{equation}

\begin{figure*}[t]
    \centering
    \includegraphics[width=\linewidth]{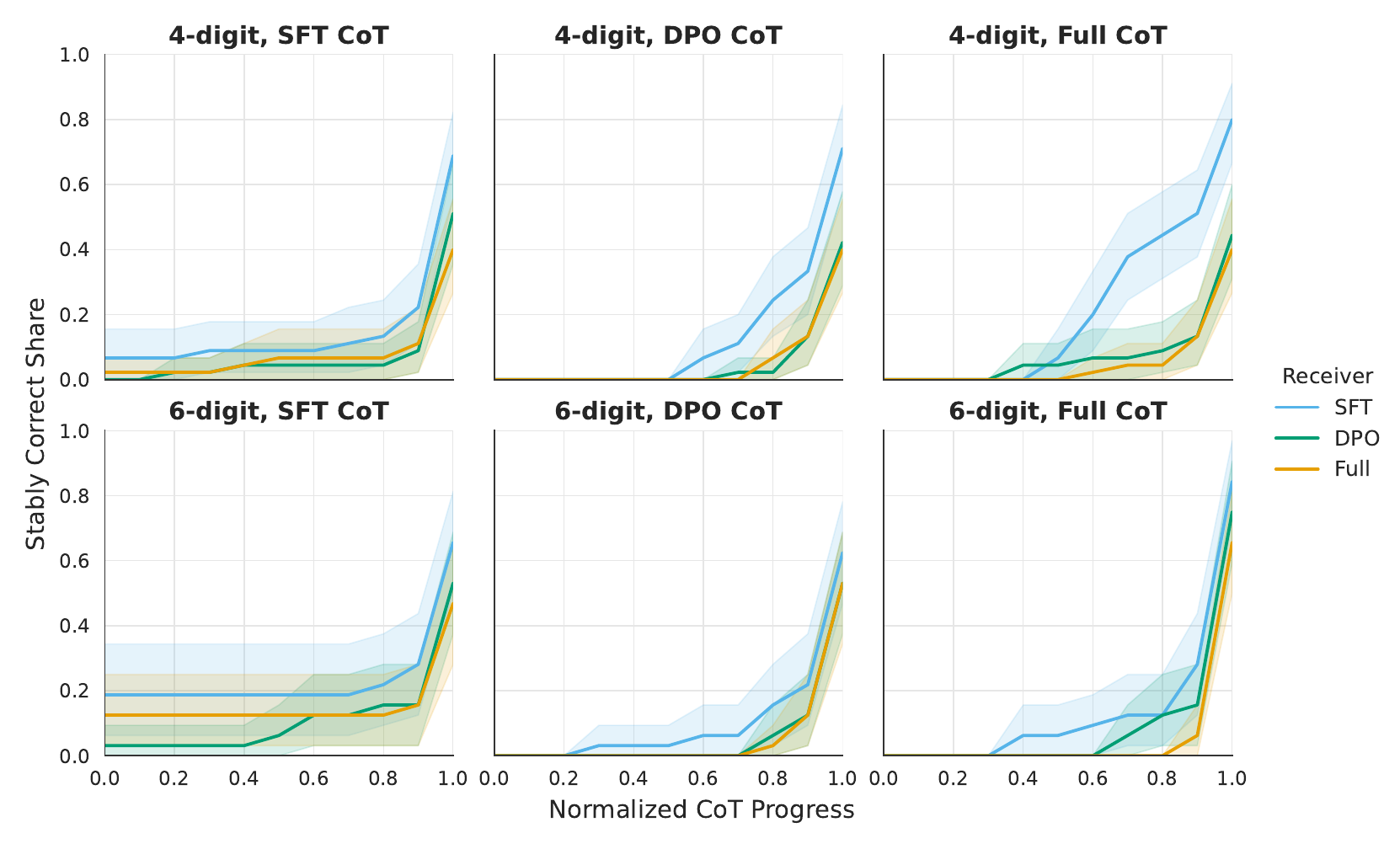}
    \caption{Cumulative observed commitment over normalized CoT progress.
    A source--receiver trajectory is committed once its directly elicited
    answer is correct at the current boundary and every later retained
    boundary. Rows show operand length, columns show the source checkpoint,
    and line colour shows the receiver checkpoint. The analysis includes all
    45 four-digit and 32 six-digit shared-source problems for each source;
    shaded regions are 95\% problem-bootstrap confidence intervals.}
    \label{fig:answer-commitment-trajectory}
\end{figure*}

If no such boundary exists, the trajectory is uncommitted. This definition is
deliberately stricter than the first correct response: an early correct answer
followed by any later incorrect answer does not count as commitment. The
cumulative curves in \Cref{fig:answer-commitment-trajectory} report the share
of trajectories with $k^{*}$ at or before each progress threshold. Unlike the
answer-emergence figure, this analysis retains source traces regardless of
whether the source checkpoint's own final answer is correct.

Commitment is concentrated near the end of the trace. At halfway through the
CoT, only 0--18.8\% of trajectories in any source--receiver cell are stably
correct; by 80\% progress, the range is still 0--44.4\%. The endpoint share
ranges from 40.0\% to 84.4\%. Thus, an answer that can sometimes be produced at the prompt is generally not stable under the subsequent reasoning trajectory.

\subsection{Decodability Is Not Commitment}

These trajectories contrast with the probe results in
\Cref{app:probes}. Incoming- and outgoing-carry information has similar
linear-probe accuracy in SFT and Full at corresponding token locations, although accuracy is lower within the CoT than at the prompt and answer. The
native answer trajectory, by contrast, changes substantially over the CoT and
usually becomes stably correct only near its end.

This distinction also qualifies findings that an eventual answer can be decoded early in a reasoning trace \citep{boppana2026reasoningtheaterdisentanglingmodel}. Early decodability does not imply that the native decoder currently prefers that answer, that the preference is robust to subsequent computation, or that the decoded feature is causally responsible for the output. In the present experiment, high prompt accuracy coexists with a subsequent fall in answerability, while stable commitment remains late. It is therefore more precise to say that answer- or carry-related information is available to a diagnostic than that the model has already settled on the answer.

The result reinforces the broader distinction between diagnostic readouts
and native answerability. One possible explanation is that reasoning
temporarily changes which partial results are favoured by the native output
mechanism, while later recomputation or verification restores a correct
answer. The trajectory measurements do not identify which process produces
the observed recovery. Future work on adaptive stopping or reasoning faithfulness should therefore compare probe readouts with native answer margins, persistence across later boundaries, and causal interventions, rather than using early probe confidence alone as evidence of commitment \citep{scalena2026commitmentboundaryprobingepiphenomenal}.

\section{Clustering Details}
\label{app:clustering}

\begin{table*}[t]
    \centering
    \small
    \setlength{\tabcolsep}{4pt}
    \renewcommand{\arraystretch}{1.08}
    \begin{tabularx}{\textwidth}{@{}lrr>{\raggedright\arraybackslash}p{0.23\textwidth}>{\raggedright\arraybackslash}p{0.28\textwidth}c@{}}
        \toprule
        Operation & Problems & Units & Units separated before K-means
        & K-means partition & Median ARI \\
        \midrule
        Addition
        & 101 & 5,598
        & Punctuation-only formatting units
        & 11 text clusters; four were merged into \textit{mixed/other}
        & 0.660 \\
        Subtraction
        & 47 & 2,783
        & Literal borrow mentions and punctuation-only formatting units
        & 10 text clusters; four were merged into \textit{mixed/other}
        & 0.695 \\
        \bottomrule
    \end{tabularx}
    \caption{Summary of the text-only clustering. Addition pools 202 SFT and
    Full reasoning traces across four- and six-digit problems. Subtraction
    pools 94 SFT and Full traces from four-digit problems. ARI is the median pairwise adjusted Rand index across ten
    single-initialization fits.}
    \label{tab:clustering-overview}
\end{table*}

\begin{table*}[t]
    \centering
    \small
    \setlength{\tabcolsep}{4pt}
    \setlength{\extrarowheight}{5pt}
    \begin{tabularx}{\textwidth}{@{}>{\raggedright\arraybackslash}p{0.19\textwidth}>{\raggedright\arraybackslash}X>{\raggedright\arraybackslash}p{0.31\textwidth}r@{}}
        \toprule
        Category & Description & Representative unit & Units \\
        \midrule
        Number decomposition
        & Extracts or aligns operand digits by place value.
        & ``6 (from 221,651)'' & 221 \\

        Incoming carry application
        & Explicitly incorporates a carry established in an earlier column.
        & ``$5 + 1$ (carry) $= 6$'' & 195 \\

        Staging headers
        & Announces the next step or column to be processed.
        & ``Step 2: Add the tens place'' & 696 \\

        Tens-place setup
        & Introduces or labels the tens-column calculation.
        & ``Tens:'' & 163 \\

        Carry tracking
        & Records an output digit and, where present, a newly produced carry.
        & ``Write down 0, carry over 1'' & 405 \\

        Step-by-step cue
        & Generic preamble introducing a worked solution.
        & ``Let's solve the addition step by step'' & 168 \\

        Repeating the problem
        & Restates the arithmetic question before solving it.
        & ``What is $1367 + 5521$?'' & 138 \\

        Problem setup
        & Short structural heading introducing the problem.
        & ``Problem:'' & 149 \\

        Formatting
        & Punctuation-only separators and answer markers.
        & ``---'' & 206 \\

        Mixed/other
        & Heterogeneous arithmetic, digit, and place-value fragments for which
          the audit did not support a single semantic label.
        & ``$9$ (from 6519) $+1$ (from 9361) $=10$'' & 3,257 \\
        \bottomrule
    \end{tabularx}
    \caption{Final addition categories. Counts are reasoning units after manual audits. The examples reproduce representative surface forms with typographical emphasis removed.}
    \label{tab:addition-cluster-categories}
\end{table*}

We group natural reasoning units by their text so that the same semantic
categories can be used across checkpoints. Addition and subtraction are fitted separately because their operation-specific vocabulary produces different partitions.

Each reasoning unit is stripped of surrounding whitespace, converted to lowercase, and represented by word-unigram and word-bigram TF--IDF features. Numeral strings, including comma-separated numerals, are replaced by a common \textsc{number} placeholder so that clusters reflect the role of a statement rather than the particular operands it contains. We retain at most 4,096 features and use sublinear term-frequency scaling. Punctuation-only units are assigned directly to a formatting category instead of being represented as zero TF--IDF vectors.

The remaining units are partitioned by K-means. The main fit uses 20 initializations. We assess sensitivity to initialization using ten additional single-initialization fits and compute the adjusted Rand index for all 45 pairs of assignments. The median pairwise adjusted Rand index is 0.660 for addition and 0.695 for subtraction. \Cref{tab:clustering-overview} summarizes the two fits.

For each initial group, we manually inspected several examples from each cluster. We manually devised labels for each group and groups without a coherent interpretation were combined into \textit{mixed/other}. We also especially assigned spans with carries (or borrows) to a specific group, to isolate the effect of these important algorithmic states. The resulting assignments were fixed before the intervention outcomes were examined.

\begin{table*}[t]
    \centering
    \small
    \setlength{\tabcolsep}{4pt}
    \setlength{\extrarowheight}{5pt}
    \begin{tabularx}{\textwidth}{@{}>{\raggedright\arraybackslash}p{0.21\textwidth}>{\raggedright\arraybackslash}X>{\raggedright\arraybackslash}p{0.30\textwidth}r@{}}
        \toprule
        Category & Description & Representative unit & Units \\
        \midrule
        Subtraction step headings
        & Announces the place-value subtraction step to be performed.
        & ``Step 3: Subtract the tens place'' & 237 \\

        Problem heading
        & Short structural heading introducing the problem.
        & ``Problem:'' & 63 \\

        Step-by-step cue
        & Generic preamble introducing a worked subtraction.
        & ``Let's solve the subtraction step by step'' & 79 \\

        Place-value headings
        & Names the current place or place-value stage.
        & ``Hundreds place:'' & 116 \\

        Result digit assembly
        & Lists place-labelled result digits or combines them into an answer.
        & ``6 (thousands), 5 (hundreds), 6 (tens), 9 (ones)'' & 136 \\

        Place-result recording
        & Records a digit in a named place after a local calculation.
        & ``Write down 0 in the tens place'' & 120 \\

        Borrow production or regrouping
        & Establishes a borrow from a higher place or converts place value.
        & ``Borrow 1 from the tens place'' & 77 \\

        Incoming borrow application
        & Applies a previous borrow by decrementing the donating place.
        & ``Reduce the hundreds digit by 1'' & 10 \\

        Lexical borrow mention
        & Other units containing a borrow-word form, including checks and
          explicit statements that borrowing is unnecessary.
        & ``No borrowing needed here'' & 327 \\

        Formatting and structure
        & Punctuation-only separators and other structural markers.
        & ``---'' & 70 \\

        Mixed/other
        & Heterogeneous calculations, answer fragments, digit adjustments,
          and presentation text without one coherent label.
        & ``$9943 - 1724$'' & 1,548 \\
        \bottomrule
    \end{tabularx}
    \caption{Final subtraction categories. Counts are reasoning units after
    the manually audited cluster merges and the two borrow refinements. The
    examples reproduce representative surface forms with typographical emphasis
    removed.}
    \label{tab:subtraction-cluster-categories}
\end{table*}

\subsection{Addition}
\label{app:clustering-addition}

The addition corpus contains 5,598 units from paired SFT and Full traces for
101 problems: 3,038 units from four-digit problems and 2,560 from six-digit
problems. After reserving the formatting group, K-means produces 11 text
clusters. Manual review retains seven of these clusters and maps four
heterogeneous clusters to \textit{mixed/other}.

Incoming-carry application was not cleanly isolated by a single K-means
cluster. We therefore use a text-only refinement that selects statements
which explicitly add a carry produced by an earlier column, including forms
such as ``plus the carried-over value'' and ``$+1$ (carry).'' Statements that
only produce or record a new carry, such as ``write down 0, carry over 1,''
remain in the carry-tracking category. This refinement assigns 195 units to
incoming carry application. \Cref{tab:addition-cluster-categories} gives the
final mutually exclusive categories used for reporting.

\subsection{Subtraction}
\label{app:clustering-subtraction}

The subtraction corpus contains 2,783 units from paired SFT and Full traces for 47 four-digit problems. Notably, we distinguish lexical mentions of borrow from spans where a borrow is actually used or produced, as this occurs frequently during reasoning. We apply the same procedure as addition otherwise when clustering, making special groups for borrow mentions. See \Cref{tab:subtraction-cluster-categories} for the final categories.

\section{Secondary Attention Suppression Results}
\label{app:secondary-attention-suppression}

\begin{figure*}[t]
    \centering
    \includegraphics[width=\textwidth]{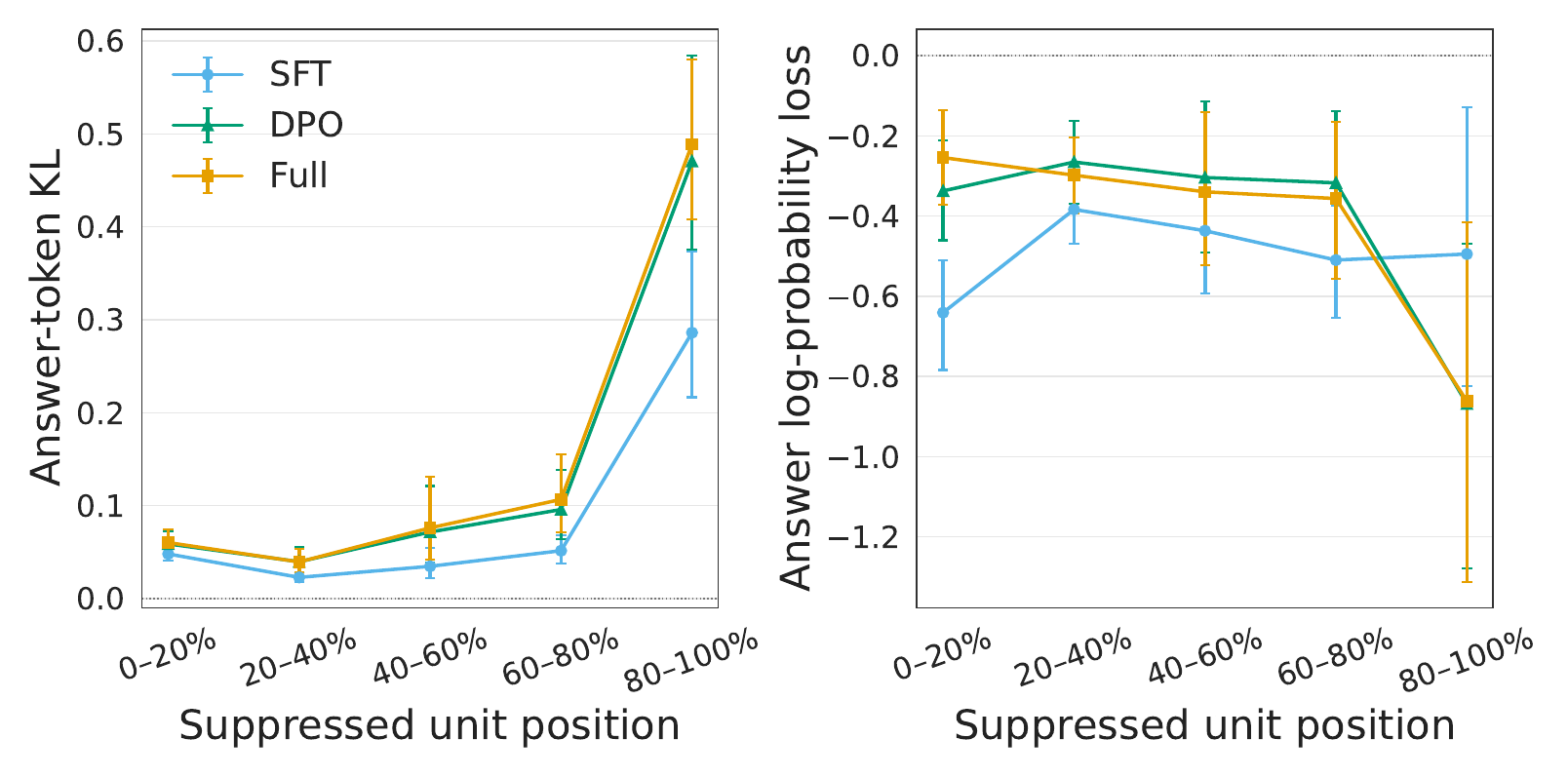}
    \caption{Effect of suppressing a reasoning unit on the final answer,
    grouped by the unit's normalized position in the CoT. The left panel shows
    the mean KL divergence over factual-answer tokens. The right panel shows
    clean minus suppressed summed log probability for those tokens; negative
    values therefore indicate that suppression increased their probability.
    Points give equal-problem-weighted means across SFT and Full source traces,
    and error bars are 95\% confidence intervals from resampling problems.}
    \label{fig:answer-dependency-by-position}
\end{figure*}

This section reports additional measurements from the attention-suppression experiments in \Cref{sec:semantic-spans}. The first three analyses use fixed, teacher-forced reasoning traces. For each source unit, later token positions are prevented from attending to that unit at every layer, while all other tokens and model weights remain unchanged. The on policy generation analysis instead allows the model to generate a new continuation while the same attention restriction remains active.

\begin{figure*}[t]
    \centering
    \includegraphics[width=\textwidth]{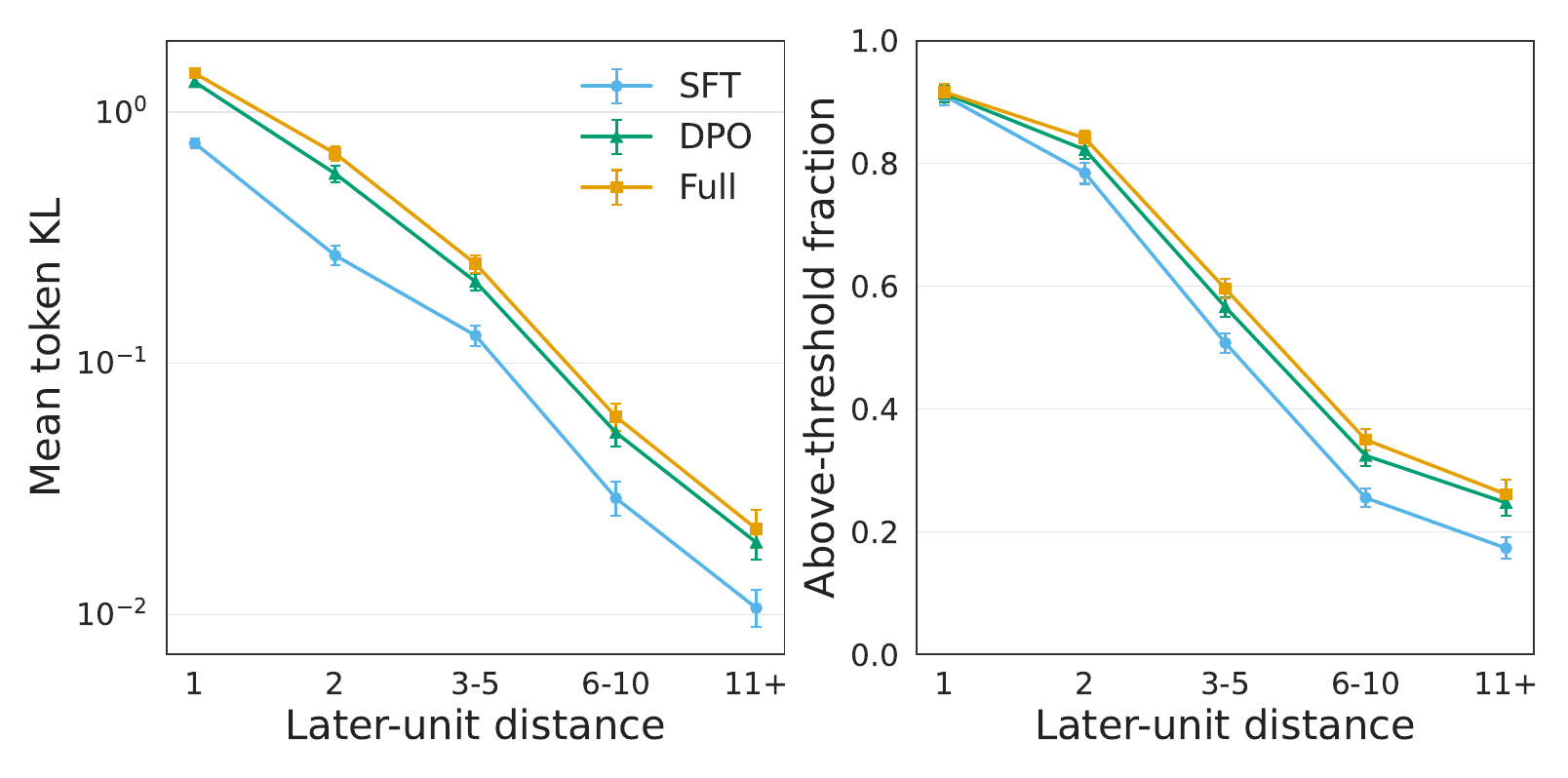}
    \caption{Dependence on an earlier reasoning unit as a function of unit
    distance. The left panel shows mean per-token KL on a logarithmic scale;
    the right panel shows the fraction of source--target pairs with mean KL of
    at least 0.01 nats. Values are averaged within each problem before being
    averaged across problems. Error bars are 95\% problem-bootstrap confidence
    intervals.}
    \label{fig:reasoning-dependency-by-distance}
\end{figure*}

\subsection{Where Suppression Affects the Final Answer}

We first ask whether a unit's position in the reasoning predicts its direct
effect on the final answer. The analysis contains paired SFT and Full traces
for 101 addition problems, comprising 202 traces and 5,598 reasoning units.
Each trace is supplied to SFT, DPO, and Full as a receiver. For a suppressed
unit $i$, we measure the mean divergence over the teacher-forced answer tokens,

\begin{equation}
    D_{\mathrm{ans}}(i)
    = \frac{1}{|A|}\sum_{t\in A}
    D_{\mathrm{KL}}\!\left(
        p_{\mathrm{clean},t}\,\|\,
        p_{\mathrm{suppressed},t}
    \right),
    \label{eq:answer-suppression-kl}
\end{equation}

where $A$ is the set of factual-answer token positions. We also report the
change in the summed log probability of those observed tokens,

\begin{equation}
    L_{\mathrm{ans}}(i)
    = \sum_{t\in A}
    \left[
        \log p_{\mathrm{clean},t}(y_t)
        - \log p_{\mathrm{suppressed},t}(y_t)
    \right].
    \label{eq:answer-suppression-logprob-loss}
\end{equation}

A positive value of $L_{\mathrm{ans}}$ means that suppression lowers the
probability of the observed answer; a negative value means that it raises that probability. The two measurements need not move together: KL divergence captures changes to the complete next-token distribution, whereas $L_{\mathrm{ans}}$ follows only the factual answer tokens.

As shown in \Cref{fig:answer-dependency-by-position}, direct changes to the
answer distribution are concentrated near the end of the reasoning. In the
first four-fifths of the trace, mean answer-token KL is at most 0.052 nats for
SFT, 0.096 for DPO, and 0.107 for Full. In the final fifth it rises to 0.286,
0.470, and 0.489 nats, respectively. The observed answer tokens themselves do
not show a corresponding probability loss: the estimates in the right panel
are negative in every position bin. Suppression can therefore alter the full
answer distribution substantially without reducing the probability assigned
to the factual token sequence.

This result does not imply that early units are unused. An early unit can
change the next reasoning step, which changes a later step, without producing
a large direct effect on the final answer under teacher forcing. The next two
analyses measure this propagation through the reasoning trace.

\subsection{Locality of Dependencies Within the CoT}

For every suppressed source unit $i$ and later unit $j$, let $T_j$ be the
token positions in the later unit. We define their directed dependence as

\begin{equation}
    D(i\!\rightarrow\!j)
    = \frac{1}{|T_j|}\sum_{t\in T_j}
    D_{\mathrm{KL}}\!\left(
        p_{\mathrm{clean},t}\,\|\,
        p_{\mathrm{suppressed},t}
    \right).
    \label{eq:reasoning-unit-dependence}
\end{equation}

Unit distance is the number of reasoning-unit boundaries between $i$ and
$j$: distance one denotes the immediately following unit. We report both the
mean value of \Cref{eq:reasoning-unit-dependence} and the fraction of directed
pairs for which it is at least 0.01 nats.

Dependence falls sharply with distance
(\Cref{fig:reasoning-dependency-by-distance}). For the immediately following
unit, the mean KL is 0.751 nats for SFT, 1.315 for DPO, and 1.424 for Full. At
a distance of 11 or more units, the corresponding values are 0.011, 0.019,
and 0.022 nats. More than 91\% of adjacent-unit pairs exceed the 0.01-nat
threshold for every receiver, compared with 17.3\%, 24.7\%, and 26.1\% of
pairs at a distance of 11 or more for SFT, DPO, and Full. Thus, the strongest
dependencies are local, although DPO and Full retain larger effects than SFT
at longer distances.

\begin{figure*}[t]
    \centering
    \includegraphics[width=\textwidth]{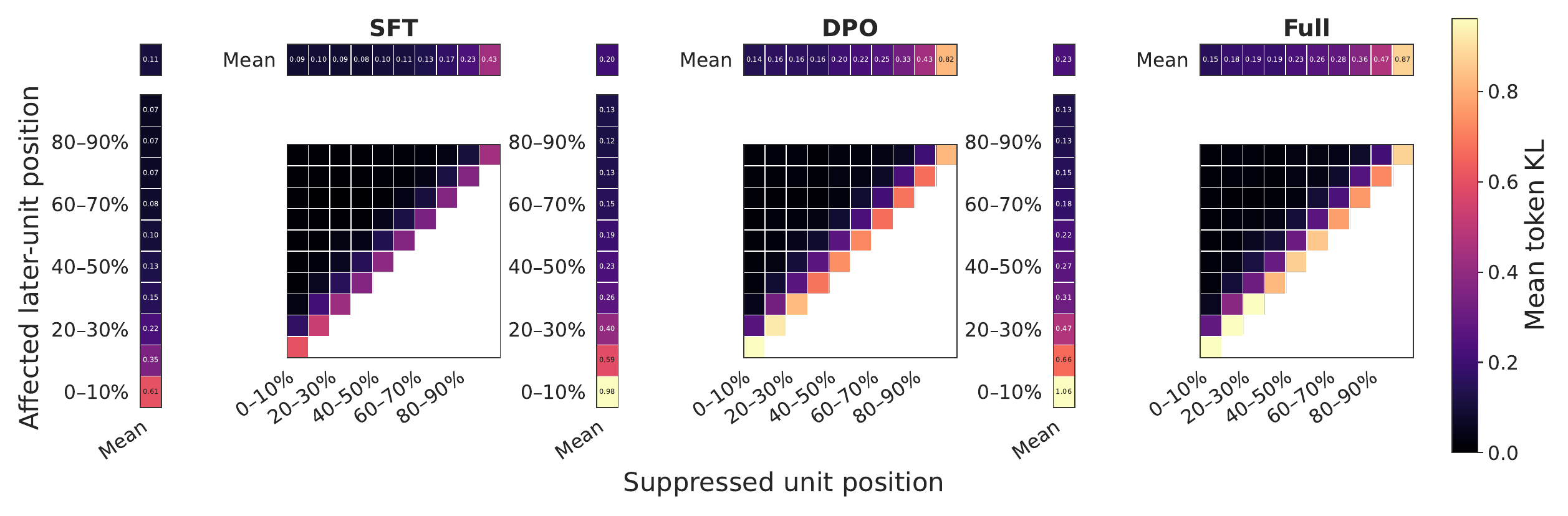}
    \caption{Map of directed dependencies between reasoning units. Source and later affected units are placed into ten equal-width bins of normalized CoT progress. Each cell gives the mean per-token KL for the observed source--target  airs in that bin; the narrow strips marked ``Mean'' summarize the populated cells along each axis.}
    \label{fig:reasoning-dependency-map}
\end{figure*}

\begin{figure*}[t]
    \centering
    \includegraphics[width=\textwidth]{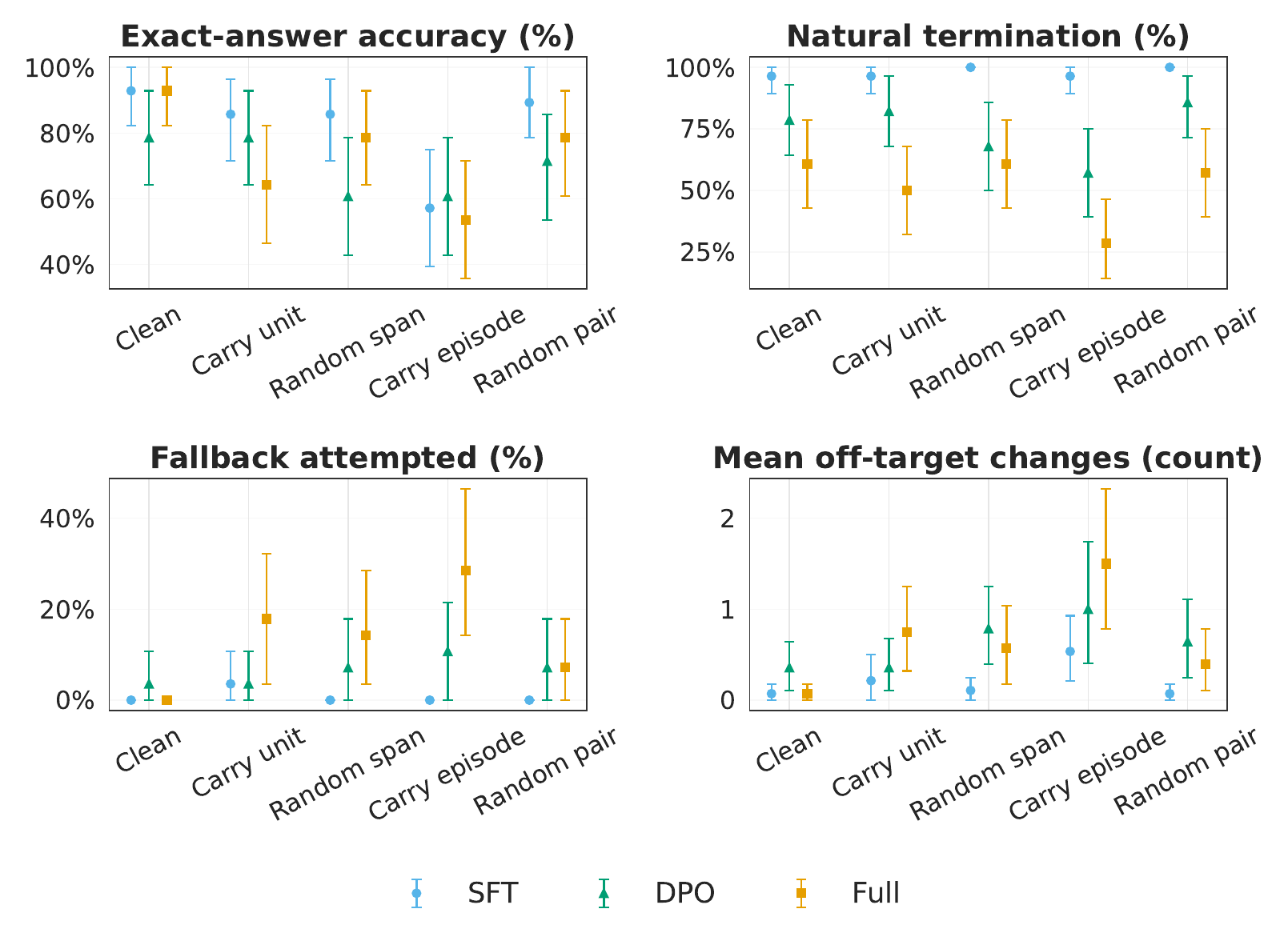}
    \caption{Secondary outcomes during on-policy generation. Conditions suppress either the incoming-carry application, a matched non-carry span, the two-unit carry episode, or a matched pair of non-carry spans. Exact-answer accuracy includes a short answer continuation used when the primary continuation contains no parseable final answer. Natural termination and fallback use are measured on the primary continuation. Off-target changes count incorrect answer digits outside the selected carry column and are computed only when an answer is parseable. Points are means and error bars are 95\% confidence intervals from resampling problems.}
    \label{fig:on-policy-secondary-outcomes}
\end{figure*}

The position maps in \Cref{fig:reasoning-dependency-map} show the same local
structure without reducing it to a single distance. The largest values form
a band directly above the source position, while distant cells are generally
darker. DPO and Full are brighter than SFT on the shared scale, especially for
sources and targets late in the trace. Together with the final-answer result,
this pattern is consistent with a chain in which early units primarily affect
nearby reasoning and answer-adjacent units have the largest direct effect on
the final distribution.

\subsection{Additional Outcomes Under On-Policy Generation}

\Cref{fig:on-policy-secondary-outcomes} shows additional outcomes for the on-policy regeneration analysis in \Cref{sec:semantic-spans}. For SFT, exact-answer accuracy falls from 92.9\% without suppression to 57.1\% under carry-episode suppression, while the matched random-pair condition reaches 89.3\%. For Full, the corresponding rates are 92.9\%, 53.6\%, and 78.6\%. DPO begins from a lower clean accuracy of 78.6\%; its carry-episode and random-pair accuracies are 60.7\% and 71.4\%.

The remaining panels show that these answer changes are not confined to the selected column. Carry episode suppression produces the largest mean number of off-target digit changes for every receiver. It also lowers natural termination and increases the use of the answer fallback most clearly for DPO and Full. These measurements indicate that removing access to both written carry statements can disrupt the continuation more broadly than the target-column outcome alone records.

\section{J-Lens Readout Calibration and Examples}
\label{app:jlens-readouts}

\subsection{Residual and readout changes}
We compare residual-stream activations across checkpoints around carry-related spans, asking where the representations differ and whether those differences are also reflected in the token predictions obtained through the J-Lens \footnote{The Jacobian Lens, or J-Lens is an interpretability technique for estimating the token-level output of intermediate layers.}. This provides a complementary view of how post-training changes the processing of carry-related text.

We compared the residual stream at three layers in each model (recall they share the same architecture, being progressive checkpoints): early (7), middle (15), and late (23). At each of these layers, we identified several tokens of interest, namely token endpoints of the previous, current, and next reasoning spans, as well as tokens before a carry value is mentioned within a span and before the resulting value is mentioned. For example, in "9 + 1 (carry-over) = 10"), we would take the token "+" as the before carry and "=" as the before result. 

We measured two metrics. First, we measured the raw \% change in the residual stream for two arbitrary activations $h_a$ and $h_b$ as:
\begin{equation}
\text{Residual change (\%)}
=
100\,
\frac{\left\lVert h_b-h_a\right\rVert_2}
     {\left\lVert h_a\right\rVert_2}.
\end{equation}

Second, we measured the Jensen-Shannon divergence between the J-Lens token readout for each model \citep{gurnee2026verbalizable}, over their entire vocabulary.

\begin{figure}
    \centering
    \includegraphics[width=\linewidth]{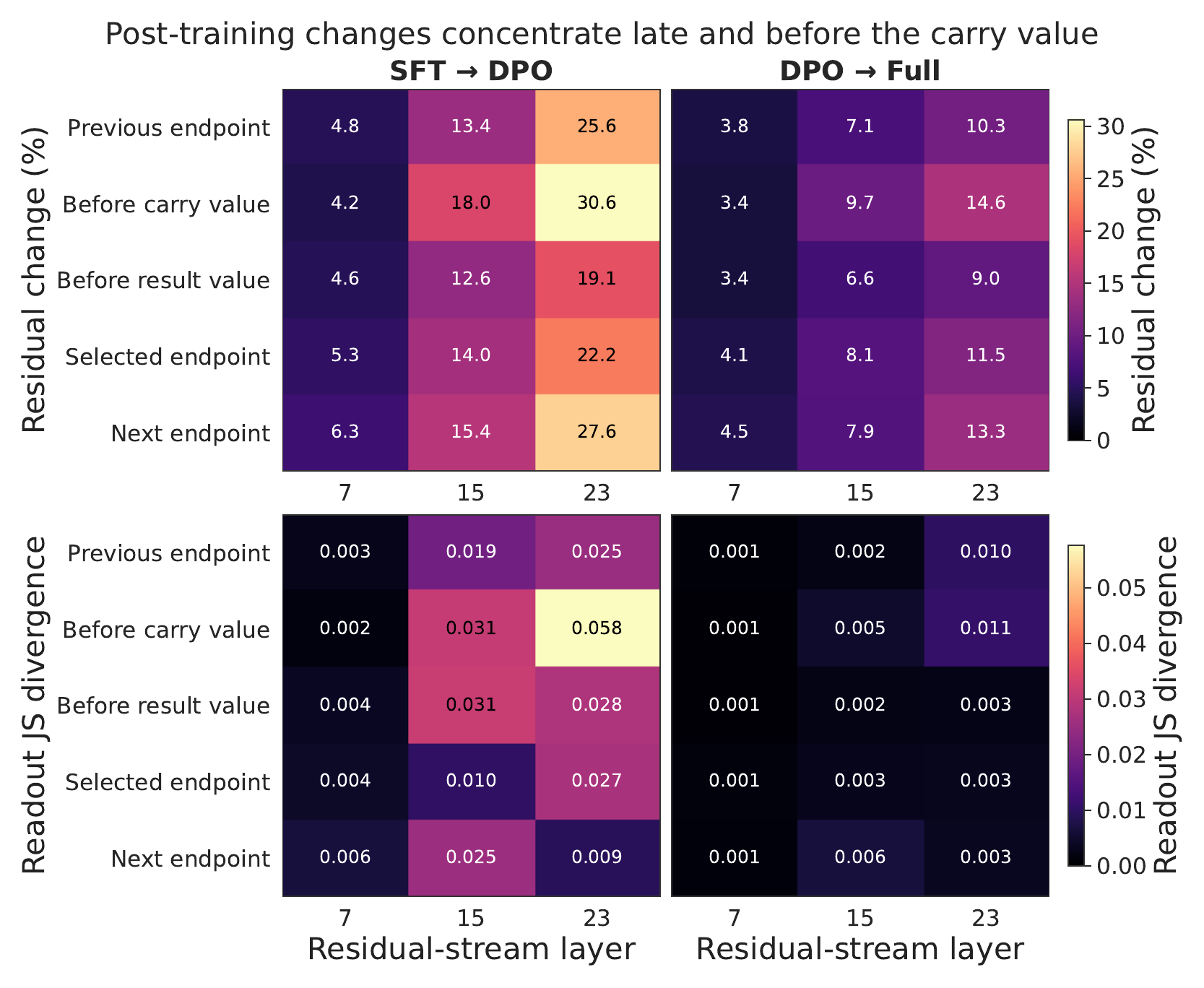}
    \caption{Analysis of the residual stream content, at  early (7), middle (15), and late (23) layers, across a carry span ((e.g. "9 + 1 (carry-over) = 10")), and at the token endpoints of the previous, current, and next reasoning spans. The change is measured between the two models shown at the top of each column. \textbf{Top}: we measure the raw \% change in residual stream value. The changes concentrate just before the carry value is mentioned. \textbf{Bottom}: using readouts from the J-Lens \citep{gurnee2026verbalizable}, we measure the Jensen-Shannon (JS) divergence between each models distribution. These results mirror those of the change in raw residual stream.}
    \label{fig:residual-readout}
\end{figure}

These two metrics, as shown in \Cref{fig:residual-readout}, tell a similar story. That is, across SFT $\rightarrow$ DPO and DPO $\rightarrow$ Full, the changes are most concentrated in the late layer and directly before the carry value. This localizes the largest measured changes within the selected carry spans. Interestingly, despite DPO performing worse than SFT or Full in the conditional accuracy comparison (c.f. \Cref{fig:acc-comparison}), the changes between SFT $\rightarrow$ DPO are much larger than DPO $\rightarrow$ Full. This pattern is compatible with the idea that later training recruits existing mechanisms \citep{venhoff2025basemodelsknowreason}, although not strong evidence of such.

\subsection{J-Lens calibration}
The Jacobian Lens (J-Lens) maps an intermediate residual-stream state into the model's final residual basis using an average input--output Jacobian \citep{gurnee2026verbalizable}. Applying the model's final normalization and output head to this transported state gives a distribution over vocabulary tokens. We use these distributions as descriptive readouts of what an intermediate state is disposed to verbalize. They are not direct measurements of the model's next-token distribution nor are they evidence that the corresponding information is causally used.

The pretrained lens used in our analysis was fitted for the \texttt{Olmo-3-7b-Base} checkpoint
on 616 generic-text sequences of 128 tokens. We apply it to layers 7, 15, and 23 of Base, SFT, DPO, and Full. These checkpoints have the same architecture and vocabulary, which permits a matched comparison, but the lens was not refitted after each post training stage. The calibration analysis below  checks how strongly the decoder chosen after the lens affects the reported token ranks.

The analysis uses 30 incoming-carry applications drawn from 23 addition problems and 26 natural reasoning traces. Sixteen units come from SFT-written traces and 14 from Full-written traces; 16 use four digit operands and 14 use six-digit operands. The units were selected for high attention-suppression importance within each operand length and source checkpoint. The sample is therefore intended for close inspection of strong carry-related cases, not as an unbiased estimate over all reasoning units.

For each unit, we read the residual stream at two locations: immediately before the incoming-carry value and immediately before the result. Each complete numeric target is represented by one token under the shared tokenizer. This gives 30 carry-value and 30 result-value locations for each checkpoint and layer.

\subsection{Effect of the Output Decoder}

Let $h_{m,l,t}$ be the residual state from checkpoint $m$, layer $l$, and
token position $t$, and let $J_l$ be the Base-trained J-Lens map for that
layer. The transported state is

\begin{equation}
    z_{m,l,t} = J_l h_{m,l,t}.
    \label{eq:jlens-transport}
\end{equation}

We decode this state in two ways. The \textit{checkpoint-native} readout applies checkpoint $m$'s final normalization and output head to $z_{m,l,t}$. The \textit{common-Base} readout instead applies the Base checkpoint's final normalization and output head to every transported state. The former retains the complete output mapping of each checkpoint. The latter holds that mapping fixed, so changes in residual representations are not mixed with changes in the final normalization or output head.

\begin{figure*}[t]
    \centering
    \includegraphics[width=\textwidth]{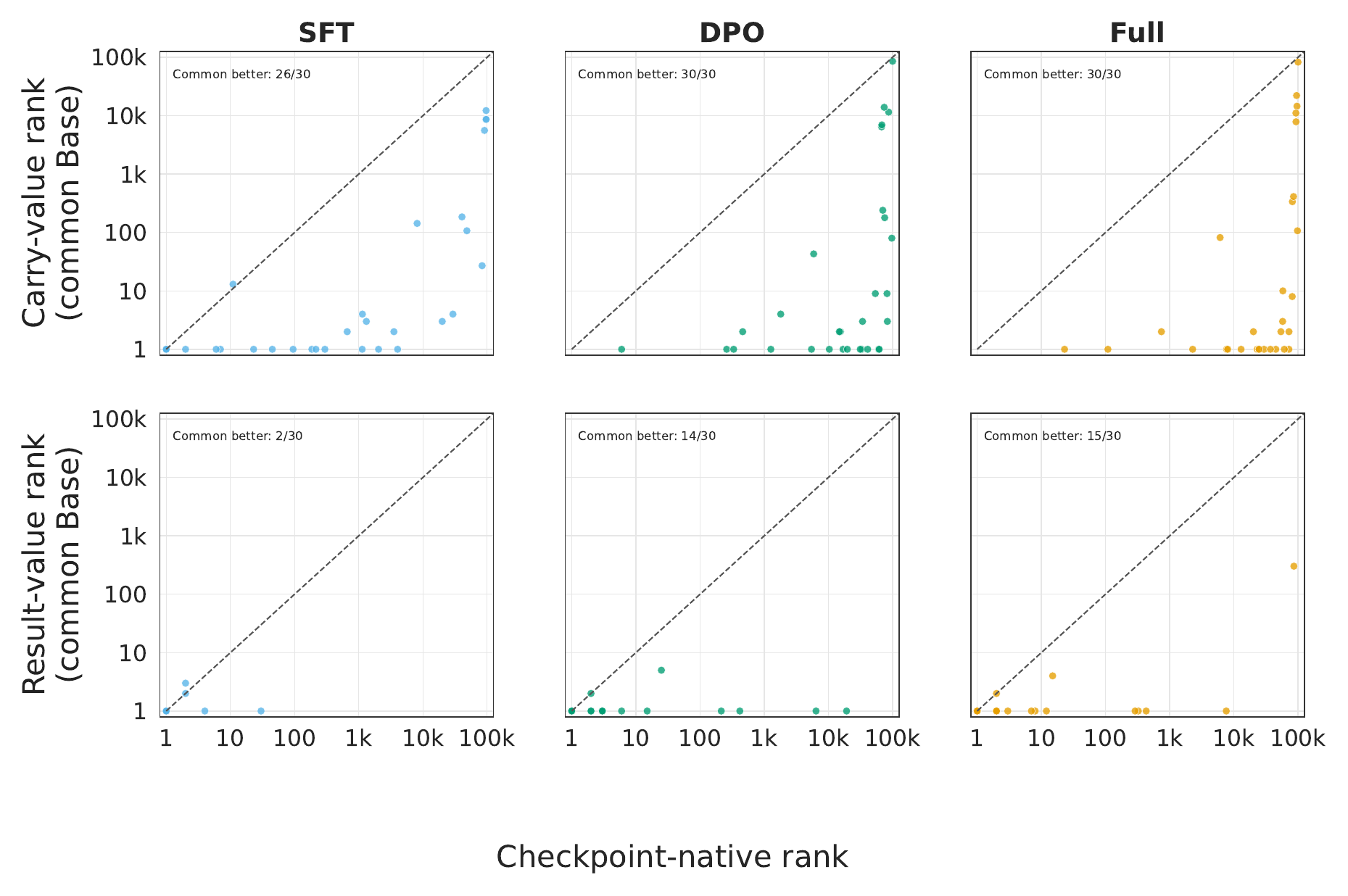}
    \caption{Comparison of checkpoint-native and common-Base J-Lens ranks at
    layer 23. Each point is one of the 30 selected carry applications. Rows
    show the known next token at the carry-value and result-value locations;
    columns show the checkpoint whose residual state is read. Both axes use a
    logarithmic scale. Points below the dashed diagonal receive a better
    (smaller) rank from the common-Base decoder. The annotation in each panel
    counts such cases; ties are not counted as improvements.}
    \label{fig:jlens-decoder-calibration}
\end{figure*}

For either readout, the vocabulary rank of the known next token is one plus the number of tokens assigned a larger logit. Rank one is therefore best. \Cref{fig:jlens-decoder-calibration} compares the two ranks at layer 23.

The decoder choice has a large effect before the carry value. The common-Base decoder gives the target token a better rank in 26 of 30 SFT cases and in all 30 DPO and Full cases. Median checkpoint-native ranks are 1,136 for SFT, 33,643 for DPO, and 49,591 for Full, whereas the corresponding common-Base medians are 1.5, 2, and 2. This mismatch is especially pronounced after the SFT stage, even though the same Base-trained transport is used throughout.

The result-value location behaves differently. Median checkpoint-native ranks are 1 for SFT, 1.5 for DPO, and 2 for Full; the common-Base median is 1 for all
three checkpoints. The common decoder is strictly better in only 2 of 30 SFT cases, 14 of 30 DPO cases, and 15 of 30 Full cases, with 27, 16, and 15 ties, respectively. Thus, the common decoder is not uniformly superior. Rather, the comparison shows that a Base-trained J-Lens and a post-trained output head can be poorly matched at some token locations, particularly immediately before a carry value. We use the common-Base readout for checkpoint comparisons because it matches the checkpoint on which the lens was fitted and keeps the decoder constant, while treating its outputs as approximate descriptions rather than ground truth.

\subsection{Illustrative Token Distributions}

\begin{figure*}[t]
    \centering
    \includegraphics[width=\textwidth]{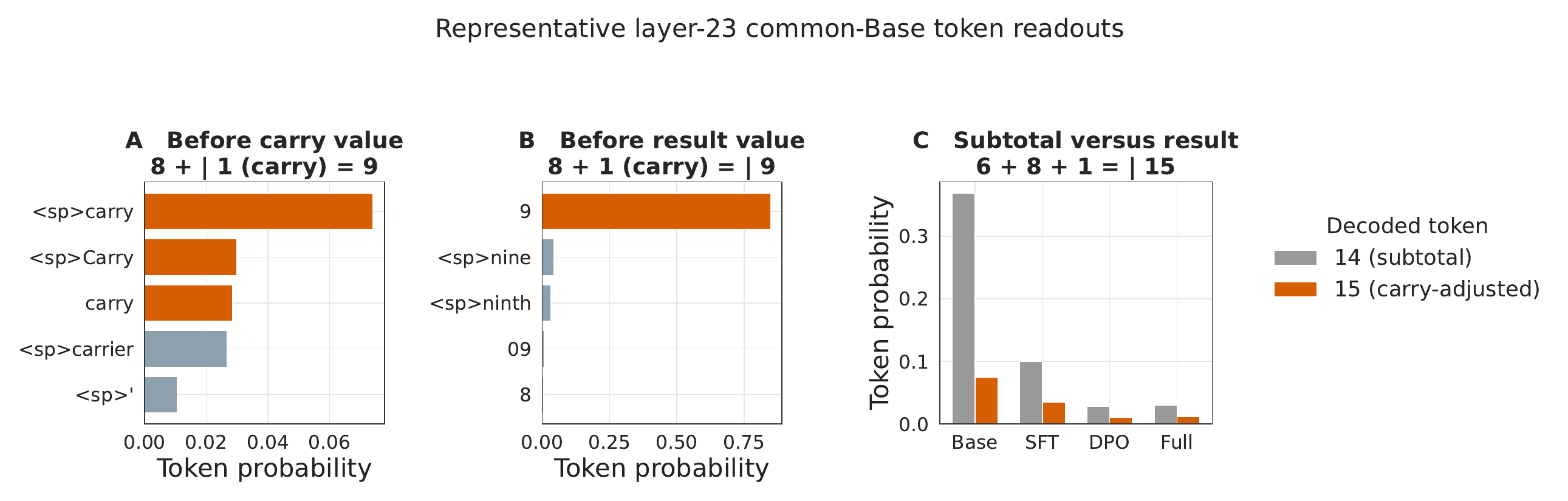}
    \caption{Illustrative layer-23 token distributions under the common-Base decoder. Panels A and B show the five highest-probability tokens from Full residual states immediately before the carry value and result value in the
    same reasoning unit; \texttt{<sp>} marks a leading space. Orange highlights carry-related tokens in Panel A and the correct numeric result in Panel B. Panel C compares the probabilities of the unadjusted subtotal 14 and the correct carry adjusted result 15 at the result location of ``$6 + 8 + 1 = 15$,'' using Base, SFT, DPO, and Full residual states with the same Base decoder. These cases are qualitative examples and are not averages over the 30-unit sample.}
    \label{fig:jlens-qualitative-readouts}
\end{figure*}

\Cref{fig:jlens-qualitative-readouts} gives two layer-23 examples under the common-Base decoder. The first two panels use the same Full residual states from the unit ``$8 + 1$ (carry) $= 9$.'' Immediately before the written carry value, the five most probable tokens are variants of the word \textit{carry}, rather than the numeric target 1. The most probable token, `` carry,'' receives probability 0.074; the target token 1 has probability 0.00056 and vocabulary rank 82. The readout is therefore aligned with the semantic role of the position but does not yet identify the exact carry value.

Immediately before the result, the distribution is much more specific. The correct target token 9 has probability 0.848 and rank one, followed by lexical
or formatting variants such as `` nine,'' `` ninth,'' and ``09.'' This contrast illustrates why a token-level readout should be examined at precise locations: within a single short arithmetic unit, it can move from a broad semantic prediction to a sharply concentrated numeric result.

The final panel shows a different failure mode. Before the result in ``$6 + 8 + 1 = 15$,'' every checkpoint's common-Base readout assigns more probability to 14, the subtotal that omits the incoming carry, than to the correct result 15. For Base the two probabilities are 0.368 and 0.075; for SFT they are 0.101 and 0.035; for DPO they are 0.028 and 0.011; and for Full they are 0.030 and 0.012. The next token in the supplied trace is nevertheless 15. This example shows that a J-Lens distribution can expose a plausible intermediate quantity without assigning its highest probability to the token that actually follows. Consequently, the readouts are most useful as descriptions of \textit{candidate} content in the residual stream.

\section{Prompt Templates}
\label{app:prompts}

We use a single user message to provide the arithmetic problem to the model. The Olmo-3 native chat template automatically prepends its default system message, which we keep:

\begin{PromptBlock}
You are a helpful function-calling AI assistant. You do not currently have access to any functions. <functions></functions>
\end{PromptBlock}

The blocks below reproduce the user-message text before chat-template
wrapping. Angle-bracketed terms denote values substituted for each example:
\texttt{<A>} and \texttt{<B>} are the two operands, and
\texttt{<A-delimited>} and \texttt{<B-delimited>} are the same kind of
operands with \texttt{|} between adjacent digits.

\subsection{Addition}

The standard-format behavioural evaluation used the following two prompts.

\paragraph{Answer Only.}
\begin{PromptBlock}
What is <A> + <B>? Give only the answer.
\end{PromptBlock}

\paragraph{Free CoT.}
\begin{PromptBlock}
What is <A> + <B>? Solve the problem step by step.
\end{PromptBlock}

The delimited-input condition inserted \texttt{|} between adjacent operand
digits while retaining the usual left-to-right answer format.

\paragraph{Delimited input, Answer Only.}
\begin{PromptBlock}
The | symbol separates digits inside a number; for example, 1|2|3 means 123. What is <A-delimited> + <B-delimited>? Give only the answer. Use standard formatting without delimiters.
\end{PromptBlock}

\paragraph{Delimited input, Free CoT.}
\begin{PromptBlock}
The | symbol separates digits inside a number; for example, 1|2|3 means 123. What is <A-delimited> + <B-delimited>? Solve the problem step by step. Use standard formatting without delimiters in your answer.
\end{PromptBlock}

The least significant digit answer condition retained conventionally written operands but required the final answer digits from least to most significant.

\paragraph{Least-Significant Digit answer, Answer Only.}
\begin{PromptBlock}
What is <A> + <B>? Give your answer from right to left with no separators; for example, if the normal answer is 6912, write 2196. This is known as the least significant digit format.
\end{PromptBlock}

\paragraph{Least-Significant Digit answer, Free CoT.}
\begin{PromptBlock}
What is <A> + <B>? Solve the problem step by step. Use least significant digit format for the final answer: write digits from right to left with no separators.
\end{PromptBlock}

\subsection{Subtraction}

All subtraction traces used standard decimal notation and the Free CoT instruction.

\begin{PromptBlock}
What is <A> - <B>? Solve the problem step by step.
\end{PromptBlock}

\subsection{Prompts Used for Continuation and Readout}

Analyses that transferred or replayed a reasoning trace retained the original standard Free CoT user message. The source checkpoint's response, or a prefix of it, was supplied as assistant text by teacher forcing. When an analysis required an answer immediately after that prefix, the following assistant-side cue was appended on a new line before scoring or decoding the answer:

\begin{PromptBlock}
Final answer:
\end{PromptBlock}

The cue was not duplicated when the supplied reasoning already ended with an equivalent final answer marker. In the on-policy attention-suppression experiment, the primary continuation began directly after the selected reasoning unit. The same cue was used only for the short fallback continuation when the primary continuation did not contain a parseable final answer.

The activation-patching analysis additionally scored the selected result digit after the following assistant-side cue, where \texttt{<column>} was replaced by the relevant place name (for example, \texttt{hundreds}):

\begin{PromptBlock}
The digit to write in the <column> column is 
\end{PromptBlock}

\subsection{Controlled Carry Continuations}

The distributed-alignment experiment used controlled four-digit scratchpads so that the incoming carry could be changed while the remaining arithmetic was held fixed. Its user message was:

\begin{PromptBlock}
What is <A> + <B>? Continue the supplied column-by-column work exactly, and end with `Final answer: <number>`.
\end{PromptBlock}

The assistant context began with the following fixed line and one templated
line for each completed column:

\begin{PromptBlock}
Let's add from right to left.
<Place>: <digit from A> + <digit from B> + carry <incoming carry> = <sum>; write <result digit> and carry <outgoing carry>.
...
\end{PromptBlock}

At the later intervention boundary, the context also ended with an explicit
statement of the carry entering the selected column:

\begin{PromptBlock}
The carry into the <column> column is <incoming carry>.
\end{PromptBlock}

The remaining column calculations and the line
\texttt{Final answer: <number> } were either supplied by teacher forcing or
generated after the intervention, depending on the measurement.

\section{Reproducibility}
\label{app:reproducibility}

\paragraph{Code.} Source code for all experiments in the project is available at \url{https://anonymous.4open.science/r/carry-trace-FBED/README.md}. 


\paragraph{Compute Resources.} All of the experiments were completed on one Nvidia A40 GPU, rented through \href{https://www.runpod.io/}{Runpod}.

\end{document}